\documentclass[journal, onecolumn, 11pt, draftcls]{IEEEtran}

\usepackage{cite}
\usepackage{enumerate}
\usepackage{graphicx}
\usepackage[cmex10]{amsmath} 
\usepackage{amssymb}
\usepackage{mathtools}
\usepackage{xspace}
\usepackage{subfigure}
\usepackage[ruled,vlined]{algorithm2e} 
\usepackage{siunitx}
\usepackage{multirow}
\usepackage{epsfig}
\usepackage{epstopdf}

\usepackage{tikz}
\usepackage{pgfplots}
\usetikzlibrary{arrows,shapes,backgrounds,plotmarks}
\pgfplotsset{compat=newest}                         
\pgfplotsset{plot coordinates/math parser=false}
\newlength\figureheight
\newlength\figurewidth

\newtheorem{theorem}{Theorem}[section]
\newtheorem{lemma}[theorem]{Lemma}
\newtheorem{definition}[theorem]{Definition}
\newtheorem{corollary}[theorem]{Corollary}
\newtheorem{proposition}[theorem]{Proposition}
\newtheorem{assumption}[theorem]{Assumption}
\newtheorem{remark}[theorem]{Remark}

\newcommand{\Z}{\mathbb{Z}}
\newcommand{\R}{\mathbb{R}}
\newcommand{\snr}{\gamma}
\newcommand{\SNR}{\Gamma}

\newcommand{\A}{\mathcal{A}}

\newcommand{\HSet}{\mathcal{H}}   
\newcommand{\B}{\mathcal{B}}      

\newcommand{\LN}{L^{\natural}}
\newcommand{\One}{\mathbf{1}}
\newcommand{\Zero}{\mathbf{0}}

\newcommand{\X}{\mathcal{X}}

\newcommand{\x}{\mathbf{x}}
\newcommand{\y}{\mathbf{y}}

\newcommand{\gv}{\mathbf{g}}

\newcommand{\ev}{\mathbf{e}}

\newcommand{\Th}{\phi}  
\newcommand{\Thv}{\mbox{\boldmath$\phi$}}  
\newcommand{\ThSet}{\Phi}         

\newcommand{\LB}{L_B} 
\newcommand{\LP}{L_P} 

\newcommand{\E}{\mathbb{E}}

\begin{document}

\title{On Monotonicity of the Optimal Transmission Policy in Cross-layer Adaptive $m$-QAM Modulation}

\author{
Ni~Ding,~\IEEEmembership{Student Member,~IEEE}, Parastoo~Sadeghi,~\IEEEmembership{Senior Member,~IEEE}, Rodney~A.~Kennedy,~\IEEEmembership{Fellow,~IEEE} \\
\thanks{
*Ni Ding is with the Research School of
Engineering, College of Engineering and Computer Science, the Australian National University (ANU), Canberra, ACT 2601, Australia (email: $\{$ni.ding$\}$@anu.edu.au).

Parastoo~Sadeghi and Rodney~A.~Kennedy are with the Research School of
Engineering, College of Engineering and Computer Science, the Australian National University (ANU), Canberra, ACT 2601, Australia (email: $\{$parastoo.sadeghi, rodney.kennedy$\}$@anu.edu.au).
}
}

\markboth{IEEE Transactions on Communications}%
{Ding \MakeLowercase{\textit{et al.}}: On Monotonicity of the Optimal Transmission Policy in Cross-layer Adaptive $m$-QAM Modulation}

\maketitle

\begin{abstract}
This paper considers a cross-layer adaptive modulation system that is modeled as a Markov decision process (MDP). We study how to utilize the monotonicity of the optimal transmission policy to relieve the computational complexity of dynamic programming (DP). In this system, a scheduler controls the bit rate of the $m$-quadrature amplitude modulation ($m$-QAM) in order to minimize the long-term losses incurred by the queue overflow in the data link layer and the transmission power consumption in the physical layer. The work is done in two steps. Firstly, we observe the $\LN$-convexity and submodularity of DP to prove that the optimal policy is always nondecreasing in queue occupancy/state and derive the sufficient condition for it to be nondecreasing in both queue and channel states. We also show that, due to the $\LN$-convexity of DP, the variation of the optimal policy in queue state is restricted by a bounded marginal effect: The increment of the optimal policy between adjacent queue states is no greater than one. Secondly, we use the monotonicity results to present two low complexity algorithms: monotonic policy iteration (MPI) based on $\LN$-convexity and discrete simultaneous perturbation stochastic approximation (DSPSA). We run experiments to show that the time complexity of MPI based on $\LN$-convexity is much lower than that of DP and the conventional MPI that is based on submodularity and DSPSA is able to adaptively track the optimal policy when the system parameters change.
\end{abstract}

\begin{IEEEkeywords}
cross-layer adaptive modulation, dynamic programming, $\LN$-convexity, Markov decision process, stochastic approximation, submodularity.
\end{IEEEkeywords}

\IEEEpeerreviewmaketitle

\section{Introduction}
\begin{figure}[tbp]
	\centering
		\centerline{\scalebox{1}{\begin{tikzpicture}

\draw [ ->] (0.9,0) -- (1.9,0);
\node at (1.2,0.25) {$f^{(t)}$};

\node at (5.1,1.2) {\small scheduler};
\draw (4.4,1.4) rectangle (5.8,1);
\draw [->,dashed] (5.1,1)--(5.1,0.35);

\node at (2.5,0.55) {\small FIFO queue};
\draw [fill=red!20] (2.8,0.35) rectangle (3.4,-0.35);
\draw (2.2,0.35) rectangle (3.4,-0.35);
\draw (2.2,0.35) -- (1.8,0.35);
\draw (2.2,-0.35) -- (1.8,-0.35);
\draw (2.5,0.35) -- (2.5,-0.35);
\draw (2.8,0.35) -- (2.8,-0.35);
\draw (3.1,0.35) -- (3.1,-0.35);

\draw [->] (3.4,0) -- (4.2,0);
\draw (4.2,0.35) rectangle (6,-0.35);
\node at (5,0.14) {\small $m$-QAM};
\node at (5.1,-0.15) {\small transmitter};
\draw [->] (6,0) -- (6.7,0);
\node at (8.2,0) {\scriptsize wireless fading channel};

\draw [dotted,color=red] (3.8,1.7) -- (3.8,-1.3);
\node at (2.2,-1) [color=red!70] {\textit{\small data link layer}};
\node at (7.2,-1) [color=blue] {\textit{\small physical layer}};

\end{tikzpicture}}}
	\caption{Cross-layer adaptive $m$-QAM system. $f^{(t)}$ denotes the number of packets arrived at data link layer at time $t$. The packet arrival process $\{f^{(t)}\}$ is random. The scheduler controls the number of bits in the QAM symbol in order to minimize the queue overflow and transmission power consumption simultaneously and in the long run.}
	\label{fig:QG}
\end{figure}
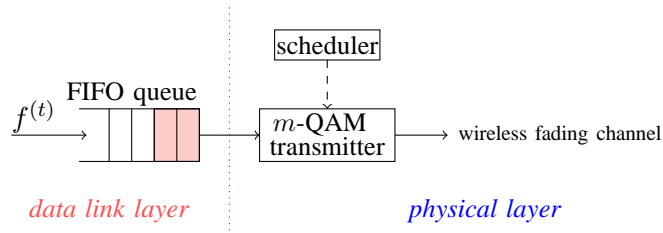

Fig.~\ref{fig:QG} shows a cross-layer adaptive $m$-quadrature amplitude modulation ($m$-QAM) system. It is assumed that packets from higher layers (e.g., application layer) arrive at the data link layer randomly. They are buffered by a first-in-first-out (FIFO) queue in the data link layer before the transmission. The physical layer adopts $m$-QAM scheme, where $m$, the constellation size, is controlled by a scheduler. In this system, $m$ determines not only the transmission rate in the physical layer but also the departure rate of the queue in the data link layer. The objective of the scheduler is to minimize the queue overflow and transmission power consumption simultaneously by considering the queue occupancy/state and channel condition/state and their expectations in the long run. The optimization problem in Fig.~\ref{fig:QG} is a cross-layer one---It incorporates the idea of adaptive modulation in the physical layer \cite{Goldsmith1997,Goldsmith1998} and the quality of service (QoS) concern associated with queueing effects in the data link layer.

There are many research works concerning cross-layer adaptive $m$-QAM system in Fig.~\ref{fig:QG}, e.g., \cite{Hoang2003,Karmokar2009,Hoang2008,Bai2008,Ding2012,Liu2005,Liu2005s}. In these works, by adopting finite-state Markov chain (FSMC) modeled wireless channel(s)\cite{Sadeghi2008}, the Markov decision process (MDP) model is proposed to formulate the dynamics (e.g., the statistics of the queue occupancy based on packet arrival probability and the variation of the channel state in FSMC) in the cross-layer adaptive $m$-QAM system and the optimal policy that minimizes the long-term losses incurred in both data link and physical layers is searched by a dynamic programming (DP) algorithm, e.g., value or policy iteration. The simulation results in these works show that scheduling across layers, instead of only one-layer, by considering the stochastic features of the system can provide good QoS and/or throughput in both data link and physical layers in the long run.

However, most of these studies focus on system model proposing and problem formulating without considering the computational complexity involved in solving the long-term optimization problems. DP is a well-known method to solve the MDP modeled optimization problems \cite{PutermanMDP1994}. However, the crucial limitation of DP is that its computation load grows drastically with the cardinalities of the state sets in MDP. This problem is called the \textit{curse of dimensionality} \cite{SuttonRL1998} and makes DP inefficient for solving high dimensional MDP problems. Take the system in Fig.~\ref{fig:QG} for example. If the number of channel states in FSMC increases, the time complexity in each iteration of DP may grow quadratically; If the system is extended to a multi-user one with MIMO (multiple-input and multiple-output) channel, the time complexity of DP may grow exponentially with both the number of users and the number of channels. In addition, DP is not suitable for real-time transmission scheduling cases either. In practical applications, we wish to design a model-free reinforcement learning algorithm that is able to quickly converge to the optimal policy and adaptively track the optimum when the system parameters change. But, DP is an off-line algorithm, i.e., running DP requires the full knowledge of MDP, and it is hard for DP to converge in real time for a large-scale MDP system when computational resources are limited. Therefore, it is worth discussing how to relieve the computational complexity of DP for the cross-layer adaptive modulation system in Fig.~\ref{fig:QG}.

On the other hand, the studies in \cite{Djonin2007,Djonin2007Struct,Huang2010,Ngo2010} show that it is possible to propose low complexity and model-free algorithm in the cross-layer optimization problem if the optimal policy is monotonic. In \cite{Djonin2007,Djonin2007Struct,Ngo2010}, a cross-layer adaptive modulation system with MIMO (multiple-input and multiple-output) channels is studied. The authors prove that the optimal transmission policy is nondecreasing in queue state/occupancy if the DP was submodular. In \cite{Djonin2007}, a modified policy iteration (MPI) algorithm is proposed based on the submodularity. It is shown that the MPI algorithm searches the optimal policy with lower complexity than DP. In \cite{Huang2010}, a multi-user adaptive $m$-QAM system is modeled by a congestion game, where the optimal randomized policy, a randomized mixture of deterministic policies, is also nondecreasing in queue state due to the submodularity. The authors propose simultaneous perturbation stochastic approximation (SPSA) algorithm for the decision maker to learn the optimal randomized policy in real time.

The main purpose of this paper is also to study how to utilize the monotonicity of the optimal transmission policy to relieve the computational complexity of DP in the cross-layer adaptive $m$-QAM system in Fig.~\ref{fig:QG}. The study is based on the MDP formulation of the $m$-QAM adaptive modulation system proposed in \cite{Hoang2003,Hoang2008}. Our work differs from the ones in \cite{Djonin2007,Djonin2007Struct,Huang2010,Ngo2010} in three aspects. Firstly, we establish the sufficient condition for the existence of a monotonic optimal transmission policy in not only the queue state but also the channel state in the MDP. Secondly, we show that the monotonicity of the optimal policy in the queue state is due to the $\LN$-convexity, a more strict property than submodularity such that the variation of the resulting optimal policy is not only nondecreasing but also restricted by a bounded marginal effect. We propose an MPI algorithm based on $\LN$-convexity and show by experiment result that its complexity is much lower than the MPI algorithm based on submodularity as proposed in \cite{Djonin2007,Djonin2007Struct}. Thirdly, the optimal policy is deterministic instead of randomized as in \cite{Huang2010}. For the purpose of learning this optimal deterministic policy in real time, we propose to use a discrete simultaneous perturbation stochastic approximation (DSPSA) algorithm based on the gradient calculation method for $\LN$-convexity in \cite{Wang2011}.

\subsection{Main Results}
The main results in this paper are listed as follows.

\begin{itemize}
    \item We prove that the optimal transmission policy is always nondecreasing in queue state due to the $\LN$-convexity of DP. It is also shown that the variation of the optimal policy in queue state is restricted by a bounded marginal effect: The increment of the optimal policy between adjacent queue states is no greater than one, i.e., if the optimal modulation scheme is $m$-QAM for a certain queue state, then the optimal modulation scheme for its adjacent queue states must be $m$-QAM, $(m+1)$-QAM or $(m-1)$-QAM.
    \item By observing the submodularity of DP, we derive the sufficient conditions for the optimal policy to be nondecreasing in both queue and channel states. We show that these conditions are satisfied if the channel experiences slow\footnote{In this paper, we assume that the fading is slow with respect to the decision duration, i.e., the normalized Doppler frequency shift, the multiplication of maximum Doppler shift and decision duration is no greater than $0.01$.} and flat fading and a proper value of the weight factor (a coefficient in the immediate cost function) is chosen.
    \item We present an MPI algorithm for searching the monotonic optimal policy based on the $\LN$-convexity of DP. It is shown that the time complexity of MPI based on $\LN$-covexity is much lower than the one based on submodularity in \cite{Djonin2007,Djonin2007Struct} and DP.
    \item We prove that the optimal transmission policy can be determined by a set of monotonic queue thresholds. For this reason, the optimal policy can be searched by solving a constrained minimization problem over queue thresholds. For solving this problem, we propose to use DSPSA algorithm, a simulation-based line search method by using augmented Lagrangian penalty method, to approximate the minimizer (the optimal queue thresholds). We run experiments to show the convergence performance of DSPSA. We show that DSPSA is able to adaptively track the optimum and optimizer when the system parameters change.
\end{itemize}

\subsection{Paper Organization}

The rest of the paper is organized as follows. In Section~\ref{sec:Model}, we describe the assumptions and MDP formulation, state the optimization objective and present DP algorithm for the adaptive $m$-QAM system in Fig.~\ref{fig:QG}. In Section~\ref{sec:Mono}, we study the existence of a monotonic optimal transmission policy in queue and channel states by observing the $\LN$-convexity and submodularity of DP. In Section~\ref{sec:MonoPoIt}, we present the MPI algorithm based on $\LN$-convexity and compare its time complexity with the one based on submodularity and DP. In Section~\ref{sec:DSPSA}, we convert DP to a discrete multivariate minimization problem with inequality constraints and show that the optimal policy can be approximated by a DSPSA algorithm.

\section{System and MDP Formulation}
\label{sec:Model}

Consider the cross-layer adaptive $m$-QAM system in Fig.~\ref{fig:QG}. Messages from higher layer are encapsulated in packets of equal length and stored in an FIFO queue in the data link layer. The output of queue is connected to an $m$-QAM transmitter in the physical layer, where the bit rate of the modulation scheme is controlled by a scheduler. The packets from higher layers (e.g., application layer) arrive at the queue in the data link layer randomly. The $m$-QAM transmitter sends packets through a wireless fading channel to the receiver. The optimization problem of the scheduler is to minimize queue overflow in the data link layer and transmission power consumption in the physical layer in the long run.

\subsection{Assumptions}
Let the decision making process be discrete, i.e., the time is divided into small intervals called \textit{decision epochs} and denoted by $t$. Each decision epoch lasts for $T_D$ seconds. Let the decision making process start from $t=0$ and go on for infinitely long time, i.e., $t\in\{0,1,\dotsc,\infty\}$. In this system, we assume the followings.

\begin{assumption}
Let $\LP$ denote the length of packet in bits. The number of storage units (in packets) in FIFO queue is $\LB<\infty$, i.e., the queue can store at most $\LB$ packets, or $\LB\LP$ bits. The newly arrived packets are dropped if there is a full queue occupancy. We call it packet loss due to the \textit{queue overflow}.
\end{assumption}

\begin{assumption}  \label{ass:Traffic}
The packet arrival process $\{f^{(t)}\}$ is \textit{i.i.d.}. $f^{(t)}\in\{0,1,\dotsc,\LB\}$ denotes the number of packets arrived at queue at $t$.
\end{assumption}

\begin{assumption} \label{ass:Action}
Let $a^{(t)}\in\{0,1,\dotsc,A_m\}$ denote the action taken by the scheduler at $t$, where the maximum action $A_m \leq \LB$. Here, $a^{(t)}=0$ denotes no transmission, and the value of $a^{(t)}$ when $a^{(t)}\neq{0}$ determines the number of bits in the QAM symbol that is transmitted by $m$-QAM transmitter at $t$, i.e., packet(s) are transmitted by $2^{a^{(t)}}$-QAM except that $a^{(t)}=0$ denotes no transmission. If $a^{(t)}\neq{0}$, the number of symbols transmitted by $m$-QAM transmitter in one decision epoch is fixed to $\LP$. For example, if $a^{(t)}=3$, $3$ packets, or $3\LP$ bits, depart from the queue. Each $3$ bits are modulate to one $2^3$-QAM symbol. The total $\LP$ $2^3$-QAM symbols are transmitted through the wireless channel. So, $a^{(t)}$ also denotes the number of packets departing from the queue at $t$ \cite{Hoang2008}. Let $T_S$ denote the symbol duration in seconds. Then, one decision epoch lasts for $T_D=\LP T_S$ seconds.
\end{assumption}

\begin{assumption} \label{ass:FSMC}
Let $\snr^{(t)}$ denote the instantaneous signal-to-noise ratio (SNR) of the wireless fading channel. $\{\snr^{(t)}\}$ is a stationary random process that is independent of $\{f^{(t)}\}$. Let the full SNR variation range of the wireless channel is partitioned into $K$ non-overlapping regions $\{[\SNR_1,\SNR_2),[\SNR_2,\SNR_3),\dotsc,[\SNR_{K},\infty)\}$, where $\SNR_1<\SNR_2<\dotsc<\SNR_K$. Denote $h^{(t)}\in\HSet=\{1,2,\dotsc,K\}$ the channel state at $t$. We say $h^{(t)}=k$ if $\snr^{(t)}\in{[\SNR_k,\SNR_{k+1})}$. The channel is modeled by an FSMC \cite{Sadeghi2008} according to the channel parameters, e.g., maximum Doppler shift, average SNR and statistics. The channel dynamics is characterized by the channel state transition probability $P_{h^{(t)}h^{(t+1)}}=\Pr(h^{(t+1)}|h^{(t)})$. The scheduler knows the value of $h^{(t)}$ to support the decision $a^{(t)}$ at each decision epoch.\footnote{The value of channel state $h^{(t)}$ can be obtained by using some channel estimation technique, e.g., \cite{Coleri2002}. We assume that the channel state does not significantly change from one decision epoch to another or when some pilot symbols are used to estimate the channel state. In this paper, we assume the perfect channel estimation and that the value of $h^{(t)}$ is known before the decision making, determining the value of $a^{(t)}$, at each decision epoch $t$. }
\end{assumption}

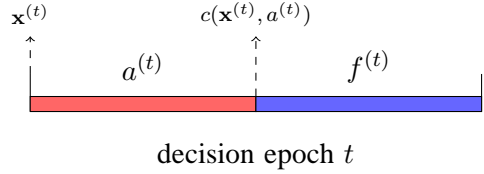
\begin{figure}[ht]
	\centering
		\centerline{\scalebox{1}{\begin{tikzpicture}

\draw (-3,0) -- (-3,0.5);
\draw (3,0) -- (3,0.5);

\draw [->,dashed] (-3,0.5) -- (-3,1);
\node at (-3,1.3) {\scriptsize $\x^{(t)}$};

\draw [fill=red!60] (-3,0) rectangle (0,0.2);
\node at (-1.5,0.6) {$a^{(t)}$};

\draw [->,dashed] (0,0.2) -- (0,1);
\node at (0,1.3) {\scriptsize $c(\x^{(t)},a^{(t)})$};

\draw [fill=blue!60] (0,0) rectangle (3,0.2);
\node at (1.5,0.6) {$f^{(t)}$};

\node at (0,-0.6) { decision epoch $t$};

\end{tikzpicture}}}
	\caption{Events happen in decision epoch $t$ in order: (1) system state $\x^{(t)}$ is observed; (2) action $a^{(t)}$ is taken; (2) immediate cost $c(\x^{(t)},a^{(t)})$ is incurred; (3) $f^{(t)}$ packet(s) arrive(s) at queue.}
	\label{fig:DE}
\end{figure}

\begin{assumption} \label{ass:Event}
The order of the events in each decision epoch is shown in Fig.~\ref{fig:DE}. At the beginning of the decision epoch $t$, the scheduler observes the system state $\x^{(t)}$ and takes an action $a^{(t)}$. A cost $c(\x^{(t)},a^{(t)})$ is immediately incurred after $a^{(t)}$. Then, $f^{(t)}$ packet(s) arrives at queue.  The definitions of $\x^{(t)}$ and $c(\x^{(t)},a^{(t)})$ will be given in Section~\ref{sec:MDP}.
\end{assumption}

\subsection{Markov Decision Process Modelling}
\label{sec:MDP}

Let $b^{(t)}\in\B=\{0,1,\dotsc,\LB\}$ be the number of packets held in the queue at decision epoch $t$. We call $b^{(t)}$ the queue state/occupancy. We define $\x^{(t)}=(b^{(t)},h^{(t)})\in\X=\B\times\HSet$ as the system state at $t$. Based on Assumptions~\ref{ass:Traffic} and \ref{ass:Event}, the variation of the queue state is governed by Lindley recursive equation \cite{Asmussen2003}
    \begin{align} \label{eq:QueNext}
        b \coloneqq \min\Big\{[b-a]^{+}+f,\LB\Big\},
    \end{align}
where $[y]^+=\max\{0,y\}$. Therefore, the queue transition probability can be worked out by the statistics of $\{f^{(t)}\}$ as
    \begin{align} \label{eq:QueTrans}
        P_{b^{(t)}b^{(t+1)}}^{a^{(t)}}&=\Pr(b^{(t+1)}|b^{(t)},a^{(t)})     \nonumber \\
           &=\begin{cases}
                \Pr(f^{(t)}=b^{(t+1)}-[b^{(t)}-a^{(t)}]^{+}) & b^{(t+1)}<\LB\\
                \sum_{l=\LB-[b^{(t)}-a^{(t)}]^+}^{\LB}\Pr(f^{(t)}=l) & b^{(t+1)}=\LB
            \end{cases}.
    \end{align}
Because of the independence of packet arrival and channel fading processes as assumed in Assumption~\ref{ass:FSMC}, the system state transition probability is given by
    \begin{align} \label{eq:Trans}
        P_{\x^{(t)}\x^{(t+1)}}^{a^{(t)}}&=\Pr(\x^{(t+1)}|\x^{(t)},a^{(t)})   \nonumber \\
                                        &=P_{b^{(t)}b^{(t+1)}}^{a^{(t)}}P_{h^{(t)}h^{(t+1)}}.
    \end{align}

Define the immediate cost $c\colon\X\times\A\mapsto\R_{+}$ as
    \begin{align} \label{eq:ImmProto}
        c(\x^{(t)},a^{(t)})&=c(b^{(t)},h^{(t)},a^{(t)})        \nonumber \\
                           &=c_q(b^{(t)},a^{(t)})+c_{tr}(h^{(t)},a^{(t)}),
    \end{align}
where $c_q$ and $c_{tr}$ quantify the costs associated with the queueing effect in the data link layer and transmission power consumption in the physical layer, respectively. We define $c_q$ as
    \begin{align} \label{eq:ImmCb}
        c_q(b^{(t)},a^{(t)})=w \E_f \bigg[ \Big[ [b^{(t)}-a^{(t)}]^{+}+f^{(t)}-\LB \Big]^{+} \bigg],
    \end{align}
where $w>0$ is a weight factor. Here, $c_q$ is proportional to the expected number of lost packets due to queue overflow. We define $c_{tr}$ as
    \begin{align} \label{eq:ImmCh1}
        c_{tr}(h^{(t)},a^{(t)})=-\frac{\ln(5\bar{P}_e)(2^{a^{(t)}}-1)}{1.5\SNR_{h^{(t)}}},
    \end{align}
where $\bar{P}_e\leq{0.2}$ is a bit error rate (BER) constraint. Here, $c_{tr}$ is an estimation of the minimum power required to transmit $a^{(t)}$ bits/symbol in channel state $h$ that will result in an average BER no greater than $\bar{P}_e$. As explained in \cite{Hoang2008}, the definition of $c_{tr}$ is based on a BER upper bound for $m$-QAM transmission derived in \cite{Foschini1983}.

Note, by using $w$, the immediate cost $c$ in \eqref{eq:ImmProto} is in fact a weighted sum of the losses incurred in data link and physical layers. The weight factor $w$ can be regarded as the priority of minimizing the cost incurred in the data link layer as opposed to that in the physical layer.

\subsection{Objective}
The optimization objective of the scheduler is to minimize the discounted sum of the immediate costs over decision epochs, which can be mathematically described as
    \begin{equation} \label{eq:objProto}
        \min \E \Big[ \sum_{t=0}^{\infty} \beta^t c(\x^{(t)},a^{(t)}) \Big| \x^{(0)}=\x \Big], \quad \forall{\x\in{\X}},
    \end{equation}
where $\beta\in[0\ 1)$ is the discount factor and $\x^{(t+1)} \sim \Pr(\cdot|\x^{(t)},a^{(t)})$. $\beta$ describes how far-sighted a decision maker is: Since $\beta$ assigns exponentially decaying weights to the immediate costs in the future, the scheduler becomes more far-sighted as $\beta\rightarrow{1}$. In addition, $\beta<1$ ensures that the limit of the infinite series is finite.

\subsection{Dynamic Programming}
Based on Assumptions~\ref{ass:Traffic} and \ref{ass:FSMC}, the MDP model in Section~\ref{sec:MDP} is stationary (time-invariant). It is proved in \cite{PutermanMDP1994} that there exists an optimal policy that is stationary and deterministic for all discounted stationary MDPs with finite state and action spaces. Therefore, by defining the expected total discounted cost under a stationary deterministic policy $\theta:\X\rightarrow\A$ as
    \begin{equation} \label{eq:V_SD}
        V_{\theta}(\x)=\E \bigg[ \sum_{t=0}^{\infty} \beta^t c(\x^{(t)},\theta(\x^{(t)})) \Big| \x^{(0)}=\x \bigg],
    \end{equation}
problem \eqref{eq:objProto} is equivalent to
    \begin{equation}  \label{eq:obj}
        \min_{\theta} V_{\theta}(\x), \quad \forall{\x}\in{\X},
    \end{equation}
Since $V_{\theta}$ can be expressed by Bellman equation \cite{Dreyfus2002}
\begin{equation} \label{eq:Bellman}
    V_{\theta}(\x)=c(\x,a)+\sum_{\x'}P_{\x\x'}^{\theta(\x)}V_{\theta}(\x'),
\end{equation}
problem~\eqref{eq:obj} can be solved by DP \cite{PutermanMDP1994}
    \begin{equation} \label{eq:DP}
        V(\x) \coloneqq \min_{a\in\A} Q(\x,a), \quad \forall{\x}\in\X,
    \end{equation}
where
    \begin{equation} \label{eq:Q}
        Q(\x,a) = c(\x,a)+\beta\sum_{\x'}P_{\x\x'}^{a}V(\x').
    \end{equation}
The optimal policy $\theta^*$ is determined by
    \begin{equation} \label{eq:OptP}
        \theta^*(\x)= \underset{a\in\A}{\operatorname{argmin}} \Big\{ c(\x,a)+\beta\sum_{\x'}P_{\x\x'}^{a}V^{(N)}(\x') \Big\}, \quad \forall{\x}\in\X,
    \end{equation}
where $N$ is the iteration index when \eqref{eq:DP} converges.\footnote{It is proved in \cite{PutermanMDP1994} that the sequence $\{V^{(n)}(\x)\}$ generated by \eqref{eq:DP} converges to $V^*(\x)$ for all $\x$, where $V^*(\x)$ is the minimum and $\theta^*(\x)=\arg\min_{a\in{\A}}\{c(\x,a)+\beta\sum_{\x'}P_{\x\x'}^{a}V^*(\x')\}$ is the minimizer of \eqref{eq:obj}. Usually, a small threshold $\epsilon^{\phantom{a}}>0$ is applied so that \eqref{eq:DP} is terminated when $\|V^{(N)}(\x)-V^{(N-1)}(\x)\|\leq{\epsilon}$ for all $\x$. In this paper, we set $\epsilon=10^{-4}$.}

Note, from \eqref{eq:Bellman} to \eqref{eq:OptP}, we drop the notation $t$ and use $\x=(b,h)$ and $\x'=(b',h')$ to denote states in the current and next decision epochs, respectively, because the MDP under consideration is stationary.

\section{Monotonic Optimal Transmission Policy}
\label{sec:Mono}
This section examines the monotonicity of the optimal transmission policy in queue and channel states. We first clarify some related definitions and theorems as follows.
\begin{definition}[Submodularity \cite{Murota2005,Hajek1985}] \label{def:submodularity}
Let $\ev_i\in{\Z^n}$ be an $n$-tuple with all zero entries except the $i$th entry being one. $f\colon\Z^n\mapsto{\R}$ is submodular if $f(\x+\ev_i)+f(\x+\ev_j)\geq{f(\x)+f(\x+\ev_i+\ev_j)}$ for all $\x\in{\Z^n}$ and $1 \leq i,j \leq n$.
\end{definition}

\begin{definition}[$\LN$-convexiy \cite{Murota2005}]  \label{def:Lconvex}
$f:\Z\mapsto{\R}$ is $\LN$-convex in $x$ if $f(x+1)+f(x-1)-2f(x)\geq{0}$ for all $x$; $f:\Z^n\mapsto{\R}$ is $\LN$-convex in $\x$ if $\psi(\x,\zeta)=f(\x-\zeta\One)$ is submodular in $(\x,\zeta)$, where $\One=(1,1,\dotsc,1)\in{\Z^n}$.
\end{definition}

In monotone comparative statics\footnotemark, it is proved that minimizing a submodular or $\LN$-convex function results in a monotonic optimal solution, which we summarize in terms of function $Q$ in the following two lemmas.
\footnotetext{Monotone comparative statics studies the situation that the optimal solution varies monotonically with the system parameters \cite{Milgrom1994}.}

\begin{lemma}  \label{lemma:SubmDP}
If $Q(\x,a)$ is submodular in $(\x,a)$, $V(\x)=\min_{a\in\A}Q(\x,a)$ is submodular in $\x$ and $a^*(\x)=\arg\min_{a\in\A}Q(\x,a)$ is nondecreasing in $\x$.
\end{lemma}
\begin{IEEEproof}
This lemma is due to the properties of submodular functions \cite{Topkis1978}: If $f(\x,\y)$ is submodular in $(\x,\y)$, $f^*(\x)=\arg\min_{\y}f(\x,\y)$ is submodular in $\x$, and $\y^*(\x)=\arg\min_{\y}f(\x,\y)$ is nondecreasing in $\x$.
\end{IEEEproof}

\begin{lemma}  \label{lemma:LconvexDP}
If $Q(\x,a)$ is $\LN$-convex in $\x$, $V(\x)=\min_{a\in\A}Q(\x,a)$ is $\LN$-convex in $\x$ and $a^*(\x)=\arg\min_{a\in\A}Q(\x,a)$ is nondecreasing in $\x$. In addition, $a^*(\x+\One)\leq{a^*(\x)+1}$ for all $\x$.
\end{lemma}
\begin{IEEEproof}
This lemma is due to the properties of $\LN$-convex functions \cite{Zipkin2008}: If $f(\x,\y)$ is $\LN$-convex in $(\x,\y)$, $f^*(\x)=\arg\min_{\y}f(\x,\y)$ is $\LN$-convex in $\x$, and $\y^*(\x)=\arg\min_{\y}f(\x,\y)$ is nondecreasing in $\x$ and $\y^*(\x+\One) \leq \y^*(\x)+1$.
\end{IEEEproof}

\begin{remark}
$\LN$-convexity differs from submodularity in that the increment of the resulting optimizer $a^*$ from $\x$ to $\x+\One$ is bounded by $1$. This is called the \textit{bounded marginal effect} \cite{Yu2013}.
\end{remark}

In this paper, the idea for proving the existence of a monotonic optimal policy is to show that the $\LN$-convexity or submodularity is preserved by minimization operation in each iteration in DP. The related results will be derived in Proposition~\ref{prop:MonoQue} in Section~\ref{sec:MonoQue} and Proposition~\ref{prop:MonoChan} in Section~\ref{sec:MonoChan}.

In the remaining context of this paper, we clarify that when we say that function $f(\x,\y)$ has some property in $\x$ we mean that $f(\x,\y)$ has this property in $\x$ for all fixed value of $\y$. For example, if $f(\x,\y)$ is nondecreasing in $\x$, then $f(\x_+,\y)\geq{f(\x_-,\y)}$ for all $\y$ if $\x_+\geq\x_-$.

\subsection{Nondecreasing Optimal Policy in Queue State}
\label{sec:MonoQue}
Based on Lemma~\ref{lemma:LconvexDP}, we show that the optimal transmission policy is always nondecreasing in queue state.
\begin{proposition} \label{prop:MonoQue}
    For $\x=(b,h)$ and $\x'=(b',h')$, if $Q(\x,a)$ is $\LN$-convex in $(b,a)$ and nondecreasing in $b$ for all $V(\x')$ that is nondecreasing and $\LN$-convex in $b'$, the optimal policy $\theta^*(\x)$ is nondecreasing in $b$, and $\theta^*(b+1,h)\leq{\theta^*(b,h)+1}$ for all $(b,h)$.
\end{proposition}
\begin{IEEEproof}
    Assume $V^{(n-1)}(\x')$ is nondecreasing and $\LN$-convex in $b'$. Then, $Q(\x,a)=c(\x,a)+\beta\sum_{\x'}P_{\x\x'}^{a}V^{(n-1)}(\x')$ is $\LN$-convex in $(b,a)$ and nondecreasing in $b$. According to Lemma~\ref{lemma:LconvexDP}, $V^{(n)}(\x)=\min_{a\in\A}Q(\x,a)$ is $\LN$-convex in $b$. Let $a^{(n-1)}(b,h)=\arg\min_{a\in\A}Q(b,h,a)$. Since
    \begin{align}
        &\quad V^{(n)}(b+1,h)-V^{(n)}(b,h)   \nonumber \\
        &=Q(b+1,h,a^{(n-1)}(b+1,h))-Q(b,h,a^{(n-1)}(b,h))   \nonumber \\
        &\geq Q(b+1,h,a^{(n-1)}(b+1,h))-Q(b,h,a^{(n-1)}(b+1,h)) \geq{0},  \nonumber
    \end{align}
    $V^{(n)}(\x)$ is also nondecreasing in $b$. Let DP starts with $V^{(0)}(\x)$ that is nondecreasing and $\LN$-convex in $b$, e.g., $V^{(0)}(\x)=0$ for all $\x$. Then, by induction, DP terminates at $N$th iteration with $c(\x,a)+\beta\sum_{\x'}P_{\x\x'}^{a}V^{(N)}(\x')$ $\LN$-convex in $(b,a)$. According to Lemma~\ref{lemma:LconvexDP}, the optimal policy $\theta^*(\x)$ determined by \eqref{eq:OptP} is nondecreasing in $b$ and $\theta^*(b+1,h)\leq{\theta^*(b,h)+1}$ for all $(b,h)$.
\end{IEEEproof}

\begin{theorem} \label{theo:MonoQue}
    The optimal policy $\theta^*(\x)$ is nondecreasing in $b$ and $\theta^*(b+1,h)\leq{\theta^*(b,h)+1}$.
\end{theorem}
\begin{IEEEproof}
    According to ~\eqref{eq:QueNext}, the queue state at the next decision epoch $b'$ can be expressed by the queue state at the current decision epoch $b$ by $b'=\min\{[b-a]^{+}+f,\LB\}$. The $Q$ function in \eqref{eq:Q} can be rewritten as
    \begin{align}
        Q(b,h,a)&=c(b,h,a)+\beta\sum_{h'}P_{hh'} \Big( \sum_{b'}P_{bb'}^{a}V(b',h') \Big)  \nonumber\\
                &=c_{tr}(h,a)+ w\E_f \Big[ [b-a]^{+}+f-\LB \Big]^{+} + \sum_{h'}P_{hh'} \E_f \Big[ V(\min\{[b-a]^{+}+f,\LB\},h') \Big].  \nonumber
    \end{align}
    Define $\varphi_o(y,f)=\Big[ [y]^{+}+f-\LB \Big]^{+}$ and $\hat{V}(y,f,h)=w \varphi_o(y,f)+ \beta V(\min\{[y]^{+}+f,\LB\},h)$. We can express $Q$ by
    \begin{equation} \label{eq:QMod}
        Q(b,h,a) = c_{tr}(h,a)+ \sum_{h'}P_{hh'} \E_f [ \hat{V}(b-a,f,h') ].
    \end{equation}
    Then, $Q$ is nondecreasing in $b$ for all $V(b',h')$ that is nondecreasing in $b'$ (see proof in Appendix~\ref{app:MonoQue2}), and $Q$ is $\LN$-convex in $(b,a)$ for all $V(b',h')$ that is $\LN$-convex in $b'$ (see proof in Appendix~\ref{app:MonoQue3}). By Proposition~\ref{prop:MonoQue}, theorem holds.
\end{IEEEproof}

\begin{remark}
Theorem~\ref{theo:MonoQue} holds unconditionally, i.e., the monotonicity of $\theta^*$ in queue state $b$ and the bounded marginal effect $\theta^*(b+1,h)\leq{\theta^*(b,h)+1}$ for all $b$ always exist regardless of the values of system parameters such as the weight factor $w$, the discount factor $\beta$, the state transition probability$P_{\x\x'}^a$.
\end{remark}

\subsection{Nondecreasing Optimal Policy in Queue and Channel States}
\label{sec:MonoChan}
Based on Lemma~\ref{lemma:SubmDP} and the results in Theorem~\ref{theo:MonoQue}, we derive the sufficient condition for the optimal policy to be nondecreasing in both queue occupancy and channel states.
\begin{proposition} \label{prop:MonoChan}
    If $Q(\x,a)$ is submodular in $(\x,a)=(b,h,a)$ for all $V(\x')$ that is nondecreasing and submodular in $\x'=(b',h')$, the optimal policy $\theta^*(\x)$ is nondecreasing in $\x=(b,h)$.
\end{proposition}
\begin{IEEEproof}
    By using Lemma~\ref{lemma:SubmDP}, this propostion can be proved by following the same induction method as in the proof of Proposition~\ref{prop:MonoQue}.
\end{IEEEproof}

\begin{theorem} \label{theo:MonoQueChan}
    If $P_{hh'}$ is first order stochastic nondecreasing\footnotemark\ in $h$ and
    \begin{align} \label{eq:MonoChanCond}
        w\leq & c_{tr}(h+1,a)+c_{tr}(h,a+1)  -  c_{tr}(h,a)-c_{tr}(h+1,a+1)
    \end{align}
    for all $(h,a)$, the optimal policy $\theta^*(\x)$ is nondecreasing in $\x=(b,h)$.
\end{theorem}
\footnotetext{See Appendix~\ref{app:1stDom} for the definition and explanation of first order stochastic dominance.}

\begin{IEEEproof}
    If $P_{hh'}$ is first order stochastic nondecreasing in $h$ and inequality \eqref{eq:MonoChanCond} holds for all $(h,a)$, we can prove that $Q$ is submodular in $(b,h,a)$ for all $V(\x')$ that is submodular in $\x'=(b',h')$ (see proof in Appendix~\ref{app:MonoQueChan}). Therefore, by Proposition~\ref{prop:MonoChan}, theorem holds.
\end{IEEEproof}

In the following two corollaries, we show that Theorem~\ref{theo:MonoQueChan} is in fact conditioned on the value of the weight factor $w$ and channel statistics.

\begin{corollary} \label{coro:MonoChan1}
If
    \begin{equation}
        w\leq{-\frac{2\ln(5\bar{P}_b)}{1.5}(\frac{1}{\SNR_h}-\frac{1}{\SNR_{h+1}})},  \nonumber
    \end{equation}
for all $h$, inequality \eqref{eq:MonoChanCond} holds.
\end{corollary}
\begin{IEEEproof}
Since
    \begin{align}
        &\quad c_{tr}(h+1,a)+c_{tr}(h,a+1) - c_{tr}(h,a)-c_{tr}(h+1,a+1)                                     \nonumber \\
        &=-\frac{2^a\ln(5\bar{P}_b)}{1.5}(\frac{1}{\SNR_h}-\frac{1}{\SNR_{h+1}})                             \nonumber \\
        &\geq -\frac{2\ln(5\bar{P}_b)}{1.5}(\frac{1}{\SNR_h}-\frac{1}{\SNR_{h+1}}),
    \end{align}
inequality \eqref{eq:MonoChanCond} holds if $w\leq{-\frac{2\ln(5\bar{P}_b)}{1.5}(\frac{1}{\SNR_h}-\frac{1}{\SNR_{h+1}})}$ holds for all $h$.
\end{IEEEproof}

The condition that $P_{hh'}$ is first order stochastic nondecreasing in $h$ in Theorem~\ref{theo:MonoQueChan} is not hard to satisfy. The following corollary shows that it holds when the channel experiences slow and flat fading with respect to the duration of decision epoch $T_D$. Here, slow means that the normalized Doppler frequency shift $f_DT_D\leq0.01$, where $f_D$ is the maximum Doppler shift.

\begin{corollary} \label{coro:MonoChan2}
If the channel experiences slow and flat fading with respect to decision duration $T_D$, the channel transition probability $P_{hh'}$ is first order stochastic nondecreasing in $h$.
\end{corollary}
\begin{IEEEproof}
Because the fading is slow and flat, the channel transitions can be worked out by level crossing rate (LCR) \cite{Sadeghi2008} and only happens between adjacent states, i.e., $h'\in\{h-1, h, h+1\}$. And, $P_{hh'}=P_{h'h}$ and $P_{hh'}\ll{P_{hh}}$ for all $h'\neq{h}$. According to Definition~\ref{def:1stDom}, for nondecreasing $u$, $P_{hh'}$ is first order stochastic nondecreasing in $h$ because
\begin{align}
        &\sum_{(h+1)'}P_{(h+1)(h+1)'}u \Big( (h+1)' \Big)-\sum_{h'}P_{hh'}u(h') \nonumber \\
        &\geq (1-2P_{h(h+1)}) \Big( u(h+1)-u(h) \Big) \geq 0,
\end{align}
where $1-2P_{h(h+1)}\geq{0}$ because $P_{hh'}\ll{P_{hh}}$ and $\sum_{h'}P_{hh'}=1$.
\end{IEEEproof}

\subsection{Examples}
We construct an adaptive $m$-QAM system as in Fig.~\ref{fig:QG}. We assume that the decision rate is $10^3$decisons/sec, i.e., the duration of each decision epoch is $T_D=10^{-3}$ second. We set queue length $\LB=15$, the maximum action $A_m=\max\A=5$ and the BER constraint $\bar{P}_e=10^{-3}$. The number of packets arrived is Poisson distributed: $f^{(t)}\sim{\text{Pois}(3)}$ for all $t$. The optimal policy $\theta^*$ is searched by DP with a discount factor $\beta=0.95$. We vary the system parameters to show the optimal transmission policies as follows.

\begin{figure}[tbp]
	\centering
        \scalebox{0.7}{\input{figures/Mono1.tex}}
	\caption{The optimal policy $\theta^*$ in a $16$-queue state $8$-channel sate cross-layer adaptive $m$-QAM system as shown in Fig.~\ref{fig:QG}, where BER constraint $\bar{P}_e=10^{-3}$, weight factor $w=1$. The channel experiences slow and flat Rayleigh fading with average SNR being $0\si{\decibel}$ and maximum Doppler shift being $10\text{Hz}$. In this system, both Theorems~\ref{theo:MonoQue} and \ref{theo:MonoQueChan} hold. $\theta^*$ is nondecreasing in queue state $b$ and channel state $h$. Since the monotonicity in $b$ is established by $\LN$-convexity, the increment of $\theta^*$ in $b$ is restricted by a bounded marginal effect, i.e., $\theta^{*}(b+1,h)\leq{\theta^{*}(b,h)+1}$ for all $(b,h)$.}
	\label{fig:Mono1}
\end{figure}

\begin{figure}[tbp]
	\centering
        \scalebox{0.7}{\input{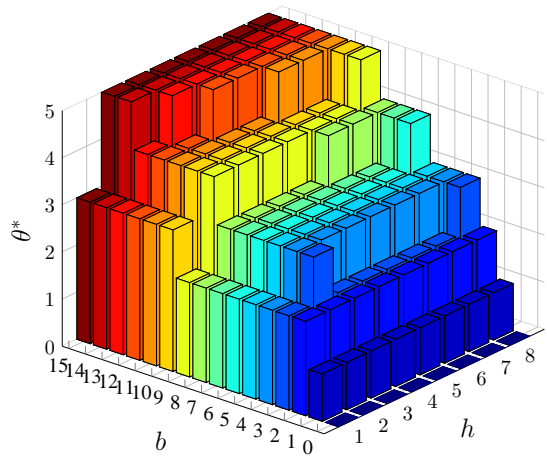}}
	\caption{The optimal policy $\theta^*$ in a $16$-queue state $8$-channel state cross-layer adaptive $m$-QAM system as shown in Fig.~\ref{fig:QG}, where BER constraint $\bar{P}_e=10^{-3}$, weight factor is $w=400$. The channel experiences slow and flat Rayleigh fading with average SNR being $0\si{\decibel}$ and maximum Doppler shift being $10\text{Hz}$. In this system, Theorem~\ref{theo:MonoQueChan} does not hold. $\theta^*$ is not nondecreasing in $h$ for all $b$, e.g, $\theta^*(b,h+1)<\theta^*(b,h)$ when $b=3$ and $h=2$. }
	\label{fig:Mono2}
\end{figure}

\begin{figure}[tbp]
	\centering
        \scalebox{0.7}{\input{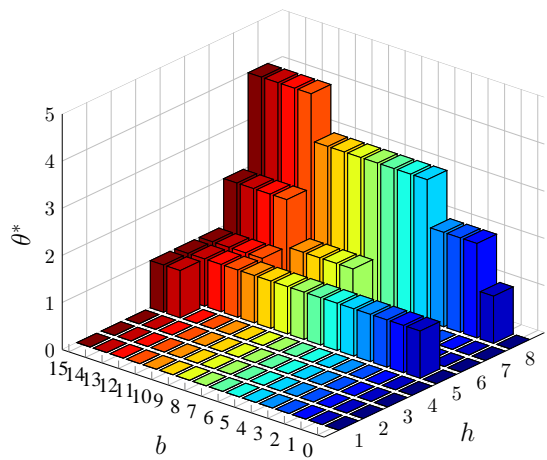}}
	\caption{The optimal policy $\theta^*$ in a $16$-queue state $8$-channel state cross-layer adaptive modulation system as shown in Fig.~\ref{fig:QG}, where BER constraint $\bar{P}_b=10^{-3}$, weight factor is $w=1$. But, the channel transition probability is not first order nondecreasing, i.e., Theorem~\ref{theo:MonoQueChan} does not hold. Therefore, $\theta^*$ is not nondecreasing in $h$ for all $b$, e.g, $\theta^*(b,h+1)<\theta^*(b,h)$ when $b=2$ and $h=5$. }
	\label{fig:Mono3}
\end{figure}

Assume the channel experiences slow and flat Rayleigh fading. Let the average SNR be $0\si{\decibel}$ and the maximum doppler shift be $10\text{Hz}$  (so that the normalized
Doppler frequency shift is $f_DT_D=0.01$). We model the channel by an $8$-state FSMC by using equiprobable SNR partition method \cite{Sadeghi2008}. We first set $w=1$. In this case, Theorem~\ref{theo:MonoQue} holds. By working out the SNR boundaries by the FSMC method described in \cite{Sadeghi2008}, it can be shown that Corollaries~\ref{coro:MonoChan1} and \ref{coro:MonoChan2} are satisfied. Therefore, Theorem~\ref{theo:MonoQueChan} also holds. As shown in Fig.~\ref{fig:Mono1}, $\theta^*$ is nondecreasing in both $b$ and $h$, and the increment of $\theta^*$ from $b$ to $b+1$ for any fixed channel state $h$ is bounded by $1$. From Fig.~\ref{fig:Mono1}, we can also see the differences between $\LN$-convexity and submodularity in terms of the resulting optimal policy: Since the monotonicity of $\theta^*$ in $b$ is due to the $\LN$-convexity, the increment of $\theta^*$ from $b$ to $b+1$ is no greater than $1$; Since the monotonicity of $\theta^*$ in $h$ is due to the submodularity instead of $\LN$-convexity, the increment of $\theta^*$ from $h$ to $h+1$ may exceed $1$, e.g., when $b=6$, the increment of $\theta^*$ from $h=7$ to $h=8$ is $2$.

We then show examples that the monotonicity of $\theta^*$ in $h$ is not guaranteed if either conditions in Theorem~\ref{theo:MonoQueChan} is breached. We first change $w$ to $400$ to breach the condition \eqref{eq:MonoChanCond}. The optimal policy is shown in Fig.~\ref{fig:Mono2}. We then set $w$ back to $1$ and change the channel transition probability as $\Pr(h'|h=7)=0$ for all $h'$ except $\Pr(h'=8|h=7)=1$ and $\Pr(h'|h=8)=0$ for all $h'$ except $\Pr(h'=1|h=8)=1$. The purpose is to satisfy \eqref{eq:MonoChanCond} but breach the stochastic dominance of $P_{hh'}$. The optimal policy is shown in Fig.~\ref{fig:Mono3}. It can be seen from Figs.~\ref{fig:Mono2} and \ref{fig:Mono3} that $\theta^*$ is not nondecreasing in $h$ for all $b$. But, since Theorem~\ref{theo:MonoQue} holds unconditionally, $\theta^*(b,h)\leq\theta^{*}(b+1,h)\leq{\theta^{*}(b,h)+1}$ for all $(b,h)$ in Figs.~\ref{fig:Mono2} and \ref{fig:Mono3}.

\section{Monotonic Policy Iteration}
\label{sec:MonoPoIt}

Consider the DP algorithm in \eqref{eq:DP}. In each iteration, a minimization operation should be done for each $\x$ in the system state space $\X$; in each minimization, the value of $Q$ is calculated for each $a$ in $\A$; and obtaining each value of $Q$ requires multiplications over all values of $\x'\in\X$. The time complexity in each iteration in DP is $O(|\X|^2|\A|)$. Since $|\X|=|\B||\HSet|$, the complexity grows quadratically if the cardinality of any tuple in the state variable increases. If the system in Fig.~\ref{fig:QG} is extended to a multi-user or multi-channel one, the time complexity of DP may grow exponentially with both the number of users and the number of channels. For example, if the wireless channel in Fig.~\ref{fig:QG} is an MIMO (multiple-input and multiple-output) one that contains $m$ subchannels, then $|\X|=|\B||\HSet|^m$, which means the time complexity of DP grows exponentially with $m$. In this and next sections, we discuss how to utilize the monotonicity results derived in Section~\ref{sec:Mono} to relieve the computational complexity of DP. For this purpose, we first propose an MPI algorithm in this section and discuss how to convert \eqref{eq:obj} to a discrete minimization optimization and apply a stochastic approximation algorithm in Section~\ref{sec:DSPSA}.

MPI is a modified DP algorithm that was first introduced in \cite{PutermanMDP1994,Djonin2007} based on the submodularity of DP. The idea is to modify the DP function in \eqref{eq:DP} as
    \begin{equation} \label{eq:SPI}
        V(\x) \coloneqq \min_{a\in\A(\x)} Q(\x,a), \quad  \forall{\x}\in\X
    \end{equation}
where $\A(\x)$ is a set or selection depending on state $\x$ and is defined as follows.

Let $\theta(\x)=\min_{a\in\A}Q(\x)$. If $\theta$ is nondecreasing in $b$ (e.g., due to the submodularity of $Q$), instead of searching the whole actions space $\A$ to get $V(\x)$, we just need to consider those actions that is no less than $\theta(b-1,h)$. Therefore, $\A(\x)$ is defined as
    \begin{equation}
        \A(\x)=\A(b,h)=\{a \colon \theta(b-1,h)\leq{a}\leq{A_m}\}. \nonumber
    \end{equation}
Note, $\A(0,h)=\A$, and \eqref{eq:SPI} should be applied in the increasing order of the value of $b$ in each iteration so that $|\A(\x)|$ is progressively reducing. MPI and DP converge at the same rate. But, the complexity in each iteration is $O(|\X|^2|\A|)$ for DP and $O(|\X|^2|\A(\x)|)$ for MPI. Since $|\A(\x)|\leq|\A|$, the computation load in MPI is less than that in DP.

\begin{figure}[tbp]
	\centering
        \scalebox{0.8}{
%
%
\begin{tikzpicture}

\begin{axis}[%
width=3.2in,
height=2.2in,
scale only axis,
xmin=2,
xmax=10,
xlabel={$|\HSet|$, the number of channel states},
ymin=0,
ymax=2100,
ylabel={Time Complexity},
legend style={at={(0.03,0.72)},anchor=south west,draw=black,fill=white,legend cell align=left}
]
\addplot [color=blue,solid,line width=2.0pt,mark=o,mark options={solid}]
  table[row sep=crcr]{2	352\\
3	528\\
4	704\\
5	880\\
6	1056\\
7	1232\\
8	1408\\
9	1584\\
10	1760\\
};
\addlegendentry{DP \cite{PutermanMDP1994}};

\addplot [color=red,solid,line width=2.0pt,mark=triangle,mark options={solid}]
  table[row sep=crcr]{2	195.356568364611\\
3	286.449591280654\\
4	376.574175824176\\
5	467.627071823204\\
6	557.759002770083\\
7	647.930555555556\\
8	731.91643454039\\
9	820.005571030641\\
10	907.156424581006\\
};
\addlegendentry{MPI based on submodularity \cite{Djonin2007}};

\addplot [color=green,solid,line width=2.0pt,mark=asterisk,mark options={solid}]
  table[row sep=crcr]{2	64\\
3	96\\
4	128\\
5	160\\
6	192\\
7	224\\
8	256\\
9	288\\
10	320\\
};
\addlegendentry{MPI based on $\LN$-convexity};

\end{axis}
\end{tikzpicture}%
}
	\caption{The time complexity of DP, MPI based on submodularity and MPI based on $\LN$-convexity in terms of the average number of calculations of $Q$ per iteration. The system settings are the same as in Fig.~\ref{fig:Mono2} except that the number of channel states in FSMC is varied from $2$ to $10$.}
	\label{fig:Tcomp}
\end{figure}
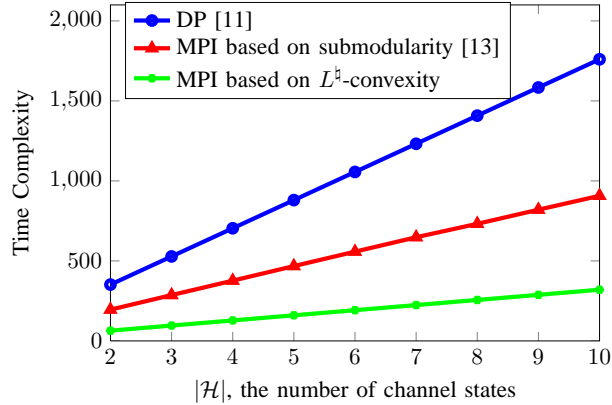

In the MDP model considered in this paper, we can show that the complexity can be further reduced. Since Theorem~\ref{theo:MonoQue} holds unconditionally, $Q$ is $\LN$-convex in $(b,a)$, and the increment of $\theta$ in $b$ is restricted by a bounded marginal effect, i.e., $\theta(b,h)$ must be either $\theta(b-1,h)$ or $\theta(b-1,h)+1$. Therefore, we can define $\A(\x)$ as
    \begin{equation}
        \A(\x)=\{\theta(b-1,h), \theta(b-1,h)+1\}. \nonumber
    \end{equation}
Therefore, the time complexity of the MPI algorithm based on the $\LN$-convexity of DP can be reduced to $O(|\X|^2)$. We use the system settings as in Fig.~\ref{fig:Mono2} and show the complexity of DP, MPI based on submodularity and MPI based on $\LN$-convexity by varying the number of channel states $|\HSet|$ in FSMC from $2$ to $10$. The results are shown in Fig.~\ref{fig:Tcomp}. In this figure, the time complexity is obtained as the number of calculations of $Q$ averaged over iterations. It can be seen that the complexity of the two MPI algorithms is less than that of DP. In addition, the complexity of the MPI algorithm based on $\LN$-convexity is much lower than the one based on submodularity.

\section{Discrete Stochastic Approximation}
\label{sec:DSPSA}

This section considers using simulation-based algorithm to relieve the complexity of DP. The idea is to convert \eqref{eq:obj} to a minimization problem over queue thresholds and use a stochastic approximation algorithm to search the optimizer. Stochastic approximation algorithms have been used in other cross-layer adaptive modulation systems before. For example, it is shown in \cite{Huang2010} that SPSA algorithm is able to learn the optimal randomized policy in an $m$-QAM congestion game. In this section, we show that stochastic approximation algorithm can also be used to search the optimal deterministic policy in the adaptive $m$-QAM system in Fig.~\ref{fig:QG}.

\subsection{Constrained Multivariate Minimization}
Based on Assumption~\ref{ass:Action}$, |\A|<|\B|$, i.e., the cardinality of the action set $\A$ is less than that of the queue state set $\B$. Since the optimal policy is always nondecreasing in queue state $b$ (Theorem~\ref{theo:MonoQue}), $\theta^{*}$ can be expressed by
    \begin{equation}   \label{eq:THPolicyH}
        \theta^{*}(\x)= \begin{cases}
                            A_m  &  \Th_{hA_m}^*\leq{b}\leq\LB   \\
                            \qquad \vdots   \\
                            1       &  \Th_{h1}^*\leq{b}<\Th_{h2}^*  \\
                            0       &  0\leq b<\Th_{h1}^*
                         \end{cases}.
    \end{equation}
Let $i\in\{1,\dotsc,A_m\}$. $\Th_{hi}^*$ is the optimal queue threshold when $\theta^*$ is switching from action $i-1$ to $i$ in channel state $h$. Define $\Thv_h=(\Th_{h1},\dotsc,\Th_{hA_m})$. $\Thv_h$ contains a set of queue thresholds that are sufficient to describe a monotonic policy for all $b$ for a certain value of $h$. Construct a queue threshold vector as $\Thv=(\Thv_1,\Thv_2,\dotsc,\Thv_{|\HSet|})$. $\Thv$ contains all queue thresholds that are sufficient to describe a policy $\theta_{\text{mono}}$ that are nondecreasing in $b$ by
    \begin{align}   \label{eq:THPolicy}
        \theta_{\text{mono}}(\x)=   \begin{cases}
                                    0  &  \{i \colon b\geq{\Th_{hi}}\}=\emptyset    \\
                                    \max\{i \colon b\geq{\Th_{hi}}\}  &  \text{otherwise}
                        \end{cases}.
    \end{align}
By doing so, \eqref{eq:obj} can be converted to a constrained multivariate minimization problem as follows.

\begin{theorem} \label{theo:objSPSA}
    The optimization problem \eqref{eq:obj} is equivalent to
        \begin{align} \label{eq:objSPSA1}
            &\qquad \min_{\Thv \in \ThSet} J(\Thv)   \nonumber \\
            &\text{s.t.}\ \Th_{hi}-\Th_{hi+1}\leq{0}, \quad \forall{h,i},
        \end{align}
    where $\ThSet=\{0,1,\dotsc,\LB+1\}^{|\HSet|\times A_m}$ and
        \begin{equation}
            J(\Thv) = \sum_{\x} \E \bigg[ \sum_{t=0}^{\infty} \beta^t c(\x^{(t)},\theta_{\text{mono}}(\x^{(t)}) \Big| \x^{(0)}=\x \bigg].
        \end{equation}
\end{theorem}
\begin{IEEEproof}
Let the set $\mathbf{\Theta}_{\text{mono}}$ contains all the deterministic stationary policies that are nondecreasing in queue state $b$. According to \eqref{eq:V_SD}, $J(\Thv)=\sum_{\x}V_{\theta_{\text{mono}}}(\x)$, where $\theta_{\text{mono}}\in\mathbf{\Theta}_{\text{mono}}$ is determined by $\Thv$ via \eqref{eq:THPolicy}. Then, \eqref{eq:objSPSA1} is in fact the problem
    \begin{equation}  \label{eq:app:obj}
        \min_{\theta_{\text{mono}}} \sum_{\x}V_{\theta_{\text{mono}}}(\x).
    \end{equation}
Since there always exists an optimal policy $\theta^*$ that is nondecreasing in $b$ (Theorem~\ref{theo:MonoQue}), $\theta^*\in\mathbf{\Theta}_{\text{mono}}$. Therefore, \eqref{eq:obj} is equivalent to \eqref{eq:app:obj}.
\end{IEEEproof}

\begin{remark}
    Since the objective function $J$ is an expectation and $\Thv$ only takes integer values, \eqref{eq:objSPSA1} is a discrete stochastic minimization problem with inequality constraints.
\end{remark}

\begin{figure}[tbp]
	\centering
        \scalebox{0.8}{\input{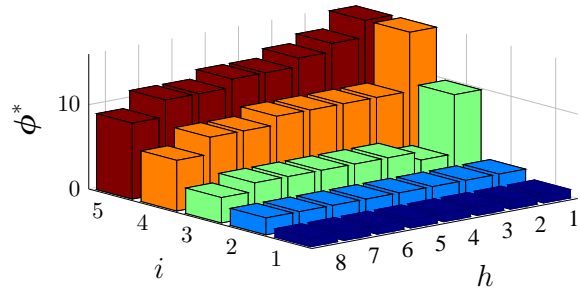}}
	\caption{The optimal queue threshold vector $\Thv^*$ extracted from the optimal transmission policy in Fig.~\ref{fig:Mono2}.}
	\label{fig:TH2}
\end{figure}

\begin{remark}
    The constrains in \eqref{eq:objSPSA1} is due to the monotonicity of $\theta_{\text{mono}}$ in $b$. Given $\theta_{\text{mono}}\in\mathbf{\Theta}_{\text{mono}}$, $\Th_{hi}$ is determined as
    \begin{equation}
        \Th_{hi}=\min\{b \colon \theta_{\text{mono}}(b,h)=i\}.
    \end{equation}
    Since $\theta_{\text{mono}}$ is nondecreasing in $b$, the queue thresholds should satisfy $\Th_{h1}\leq{\Th_{h2}}\leq\dotsc\leq{\Th_{hA_m}}$. See examples in Figs.~\ref{fig:TH1} and \ref{fig:TH2}.
\end{remark}

\begin{figure}[tbp]
	\centering
        \scalebox{0.8}{\input{figures/TH1.tex}}
	\caption{The optimal queue threshold vector $\Thv^*$ extracted from the optimal transmission policy in Fig.~\ref{fig:Mono1}.}
	\label{fig:TH1}
\end{figure}

\subsection{Discrete Simultaneous Perturbation Stochastic Approximation}
Consider using stochastic approximation algorithm to solve problem \eqref{eq:objSPSA1}. We present a DSPSA algorithm in Algorithm 1. This algorithm was first proposed in \cite{Wang2008SA}. It uses gradient based line search iterations and augmented Lagrangian method\footnote{Augmented Lagrangian is a combination of penalty and Lagrangian methods for solving constrained minimization problems. It was suggested in \cite{Nocedal2008} to prevent the situation when the penalty coefficient goes to infinity with the iteration index as in quadratic penalty method. For more details on augmented Lagrangian, we refer the reader to \cite{Nocedal2008}.} to solve an inequality constrained stochastic minimization problem. It produces an estimation sequence of the minimizer $\{\tilde{\Thv}^{(n)}\}$ with $\tilde{\Thv}^{(n)}\in\tilde{\ThSet}=[0,\LB+1]^{\HSet\times A_m}$. In Algorithm 1, $\upsilon_{hi}$ is the constraint function in \eqref{eq:objSPSA1}, i.e.,
\begin{equation}
    \upsilon_{hi}(\Thv)=\Th_{hi}-\Th_{hi+1},
\end{equation}
and $\Pi_{\tilde{\ThSet}}(\tilde{\Thv})$ is a projection function that returns a closest integer point (by Euclidean distance) in $\ThSet$ to $\tilde{\Thv}$. The implementation details of Algorithm 1 are described as follows.

	\begin{algorithm} [t]
	\label{algo:DSPSA}
	\small
	\SetAlgoLined
	\SetKwInOut{Input}{input}\SetKwInOut{Output}{output}
	\SetKwFor{For}{for}{do}{endfor}
	\Input{initial guess $\tilde{\Thv}^{(0)}$ (a $D$-tuple with $D=|\HSet|{A_m}$), total number of iterations $N$, step size parameters $A$, $B$, $\alpha_1$ and $\alpha_2$ and the penalty coefficient $R$}
	\Output{$\Pi_{\tilde{\ThSet}}(\tilde{\Thv}^{(N)})$}
	\BlankLine
	\Begin{
        set Lagragian multiplier $\lambda_{hi}^{(0)}=0$ for all $h$ and $i$\;
        \For {n=1 \emph{\KwTo} N} {
            $a^{(n)}=\frac{A}{(B+n)^{\alpha_1}}$\;
            $r^{(n)}=Rn^{\alpha_2}$\;
            obtain $\gv$ at $\tilde{\Thv}^{(n-1)}$ by using simulated objective function $\hat{J}$\;
            update estimation by
            \begin{equation}
                \tilde{\Thv}^{(n)}= \tilde{\Thv}^{(n-1)}-a^{(n)} \Big( \mathbf{g}(\tilde{\Thv}^{(n-1)})+ \sum_{h}\sum_{i} \max\{0,\lambda_{hi}^{(n-1)}+r^{(n)}\upsilon_{hi}(\tilde{\Thv}^{(n)})\} \nabla{\upsilon_{hi}(\tilde{\Thv}^{(n)})} \Big) ;   \nonumber
            \end{equation}
            update Lagrangian multiplier by
            \begin{equation}
                \lambda_{hi}^{(n)}=\max\Big\{0,\lambda_{hi}^{(n-1)}+r^{(n)}\upsilon_{hi}(\Thv^{(n)})\Big\} \nonumber
            \end{equation}
            for all $h$ and $i$\;
        }
	}
	\caption{DSPSA \cite{Wang2008SA}}
	\end{algorithm}

\subsubsection{Obtain $\mathbf{g}$}
Since \eqref{eq:objSPSA1} is a discrete optimization problem, we use the gradient calculation method based on discrete midpoint convexity in \cite{Wang2011}. The method is to generate $\mathbf{\Delta}=(\Delta_1,\dotsc,\Delta_D)$ with each tuple $\Delta_d\in\{-1,1\}$ being independent Bernoulli random variables with probability $0.5$. The $d$th entry of $\gv(\tilde{\Thv}^{(n)})$ is obtained by
\begin{align}
    g_d(\tilde{\Thv}^{(n)})=& \bigg( \hat{J} \Big( \lfloor\tilde{\Thv}^{(n)}\rfloor+\frac{\One+\mathbf{\Delta}}{2} \Big) - \hat{J} \Big( \lfloor\tilde{\Thv}^{(n)}\rfloor+\frac{\One-\mathbf{\Delta}}{2} \Big) \bigg) \Delta_d^{-1}.
\end{align}

\subsubsection{Obtaining $\hat{J}$}
$\hat{J}$ is the noisy measurement of the objective function $J$. The method of obtaining $\hat{J}(\tilde{\Thv})$ is to simulate the sequence $\{\x^{(t)}\}$. Here, $\{\x^{(t)}\}$ is governed by the Markov chain with the state transition probability being $\Pr(\x^{(t+1)}|\x^{(t)})=P_{\x^{(t)}\x^{(t+1)}}^{\theta_{\text{mono}}(\x^{(t)})}$. $\theta_{\text{mono}}(\x)$ is determined by $\tilde{\Thv}$ via \eqref{eq:THPolicy}. We obtain $\hat{J}$ as
\begin{align} \label{eq:SimJ}
\hat{J}(\tilde{\Thv})=\sum_{\x^{(0)}\in\X}\sum_{t=0}^{T}\beta^{t}c(\x^{(t)},\theta_{\text{mono}}(\x^{(t)})).
\end{align}
$T$ is the simulation length and depends on $\beta$, i.e., the simulation stops until the increments over several successive decision epochs are blow a small threshold ($10^{-4}$).

\subsubsection{Obtaining $\nabla{\upsilon_{hi}(\tilde{\Thv}^{(n)})}$}
$\nabla{\upsilon_{hi}(\tilde{\Thv}^{(n)})}$ is the gradient of the constraint function $\upsilon_{hi}$ at $\tilde{\Thv}^{(n)}$. Since $\upsilon_{hi}$ is linear. $\nabla{\upsilon_{hi}(\tilde{\Thv}^{(n)})}$ is simply the coefficients in $\upsilon_{hi}$.

\subsubsection{Step Size Parameters and Penalty Coefficient}
The step size parameters, $A$, $B$, $\alpha_1$ and $\alpha_2$, and the penalty coefficient $R$ in Algorithm 1 are crucial for the convergence performance of DSPSA algorithms. In this paper, we adopt the method of choosing $A$, $B$, $\alpha_1$, $\alpha_2$ and $R$ suggested in \cite{SpallPar1998,Wang2008SA}\footnote{The authors in \cite{SpallPar1998} presented an implementation guide for the designers to choose the step size parameters when applying simultaneous perturbation stochastic approximation (SPSA) method for practical problems. The experiments in the subsequent works, e.g., \cite{Wang2008SA}, proved that this method could provide good convergence performance for SPSA algorithms. }: $A=0.015$ $B=100$, $\alpha_1=0.602$, $\alpha_2=0.1$ and $R=10$. DSPSA always starts with $\tilde{\Thv}^{(0)}=\Zero$.

\begin{figure*}[t]
	\centering
        \subfigure[{$J([\tilde{\Thv}^{(n)}])$}, the value of the objective function, at the $n$th estimation]{\includegraphics[height=5cm,width=8cm]{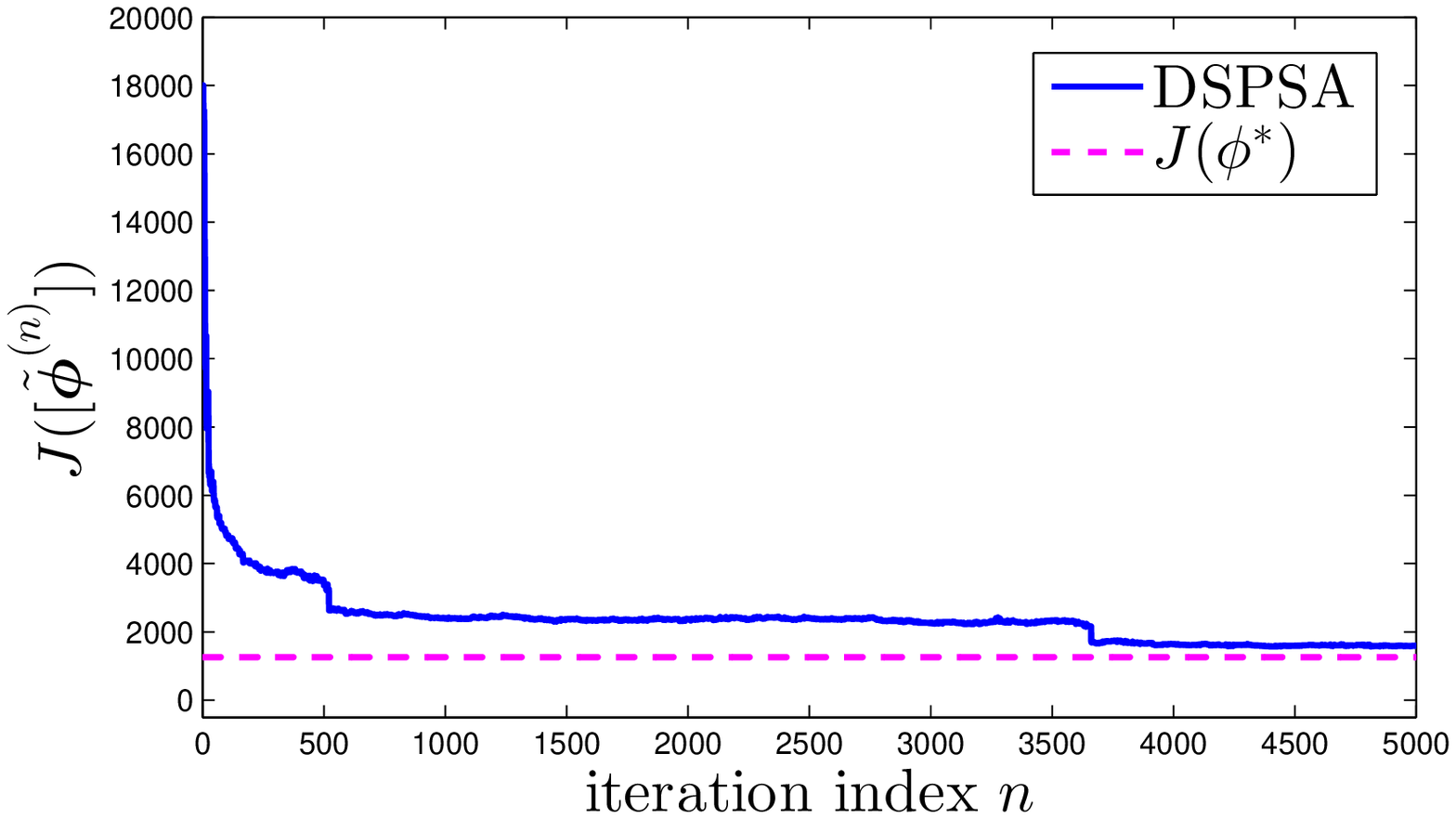}}
        \subfigure[$\frac{\|\tilde{\Thv}^{(n)}-\Thv^*\|}{\|\tilde{\Thv}^{(0)}-\Thv^*\|}$, the normalized error, at the $n$th estimation]{\includegraphics[height=5cm,width=8cm]{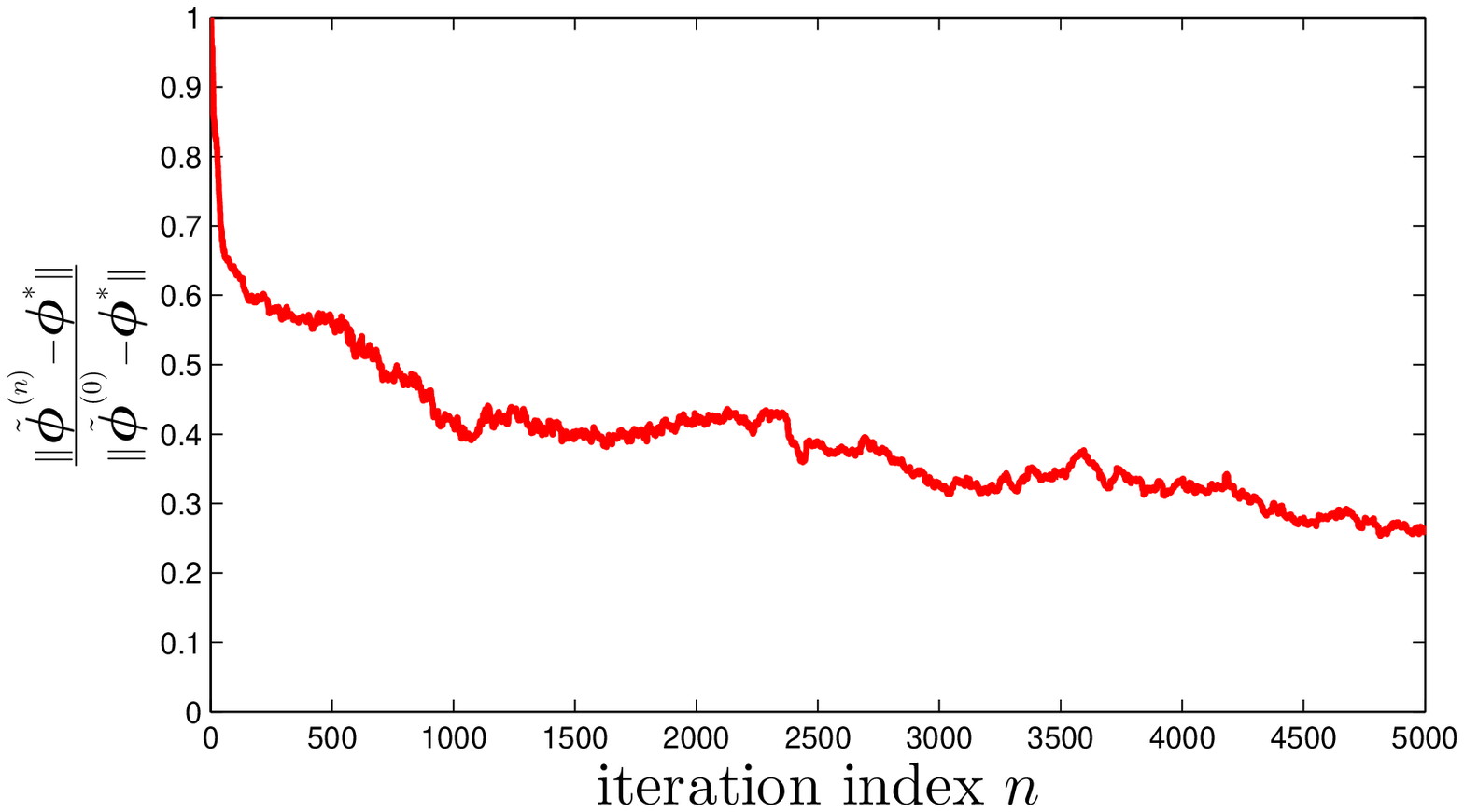}}
	\caption{Convergence performance of DSPSA when $\LB=15$, $A_m=5$, $f^{(t)}\sim{\text{Pois}(3)}$ and $\bar{P}_e=10^{-3}$. The channel is Rayleigh fading with average SNR being $0\si{\decibel}$ and maximum Doppler shift being $10\text{Hz}$. It is modeled by a $8$-state FSMC. The weight factor is $w=100$.}
	\label{fig:SPSA2}
\end{figure*}

\begin{figure*}[t]
	\centering
        \subfigure[{$J([\tilde{\Thv}^{(n)}])$}, the value of the objective function, at the $n$th estimation]{\includegraphics[height=5cm,width=8cm]{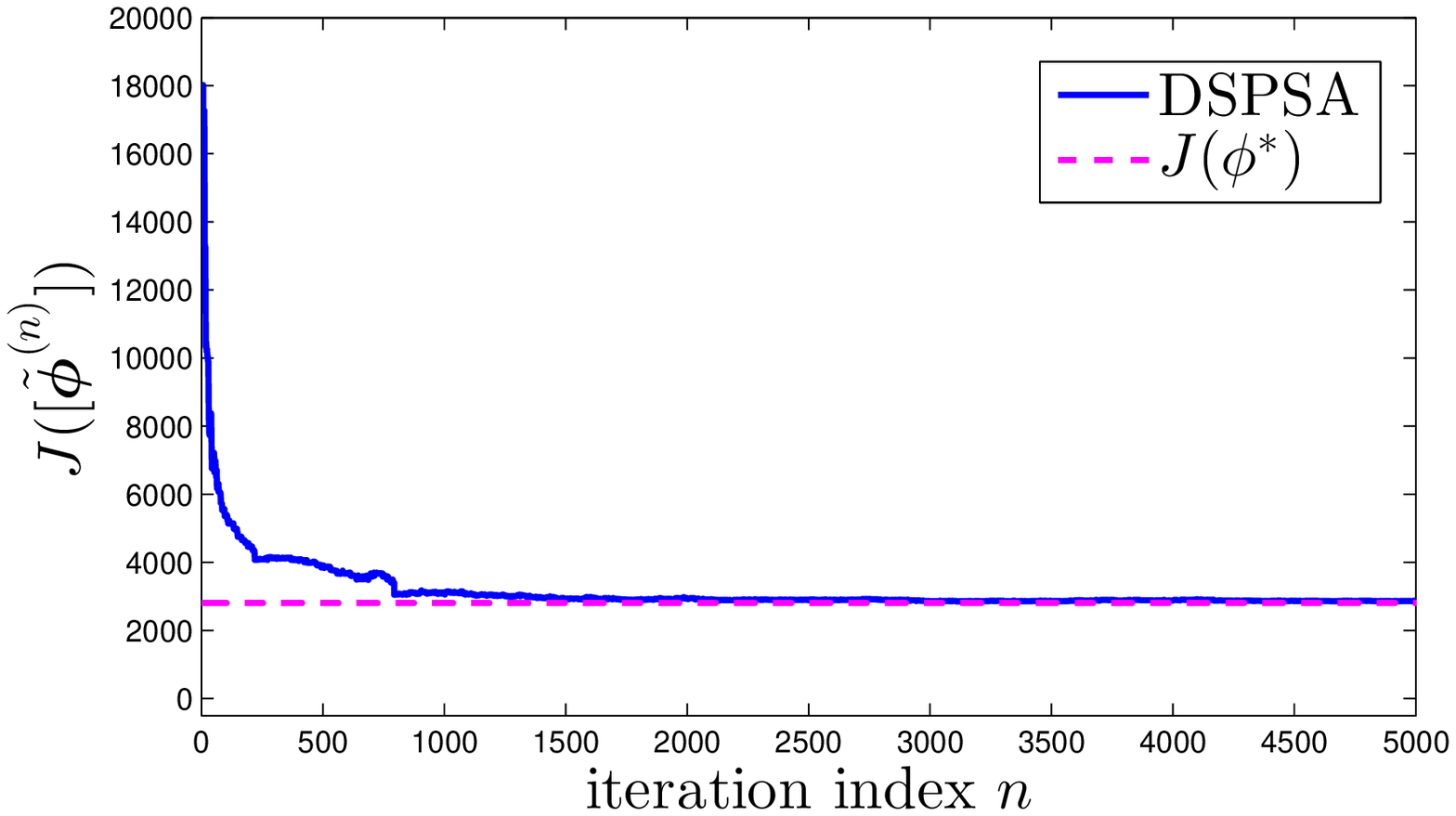}}
        \subfigure[$\frac{\|\tilde{\Thv}^{(n)}-\Thv^*\|}{\|\tilde{\Thv}^{(0)}-\Thv^*\|}$, the normalized error, at the $n$th estimation]{\includegraphics[height=5cm,width=8cm]{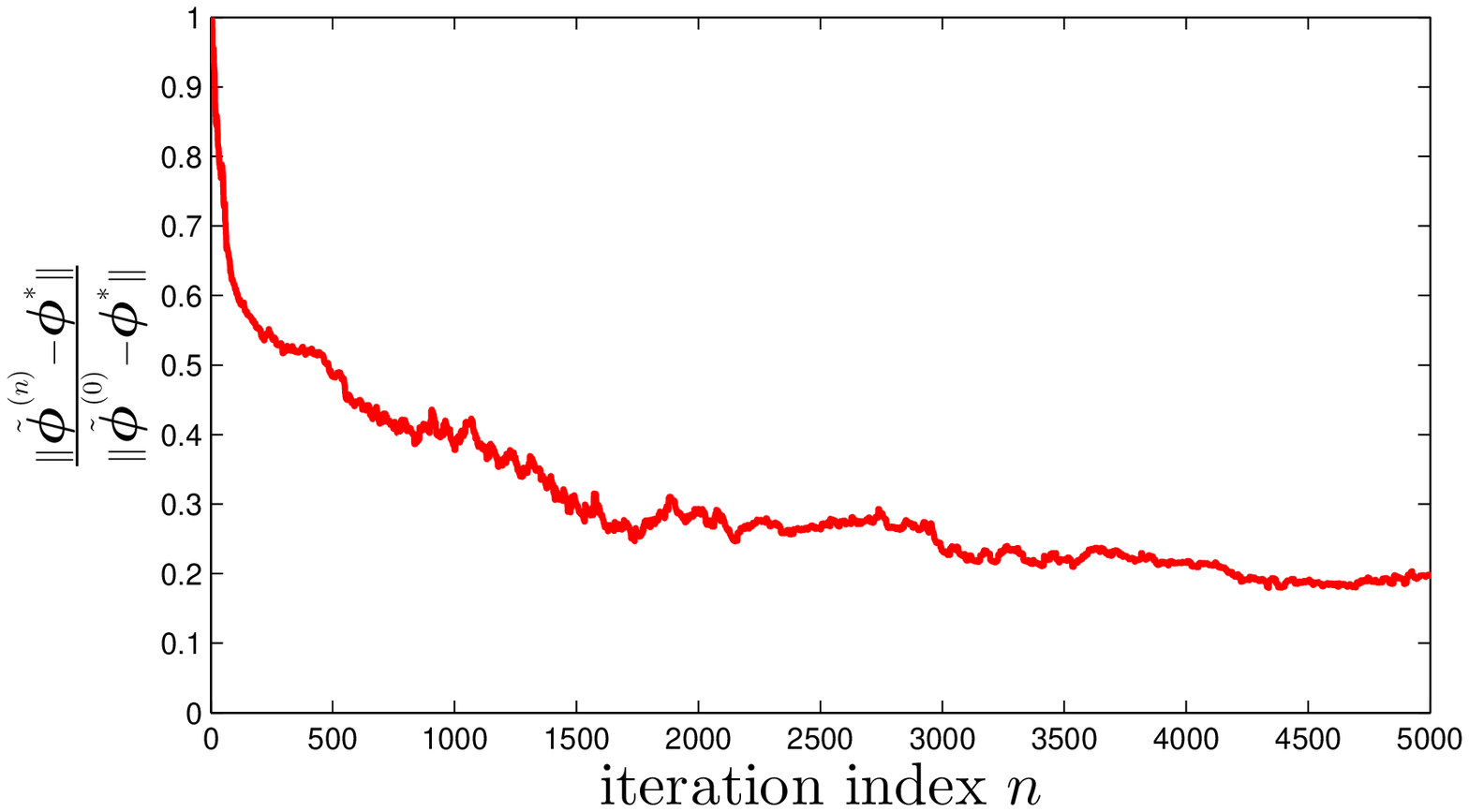}}
	\caption{Convergence performance of DSPSA when the parameters are the same as in Fig.~\ref{fig:SPSA2} except that $w=400$.}
	\label{fig:SPSA3}
\end{figure*}

\begin{figure*}[t]
	\centering
        \subfigure[{$J([\tilde{\Thv}^{(n)}])$}, the value of the objective function, at the $n$th estimation]{\includegraphics[height=5cm,width=8cm]{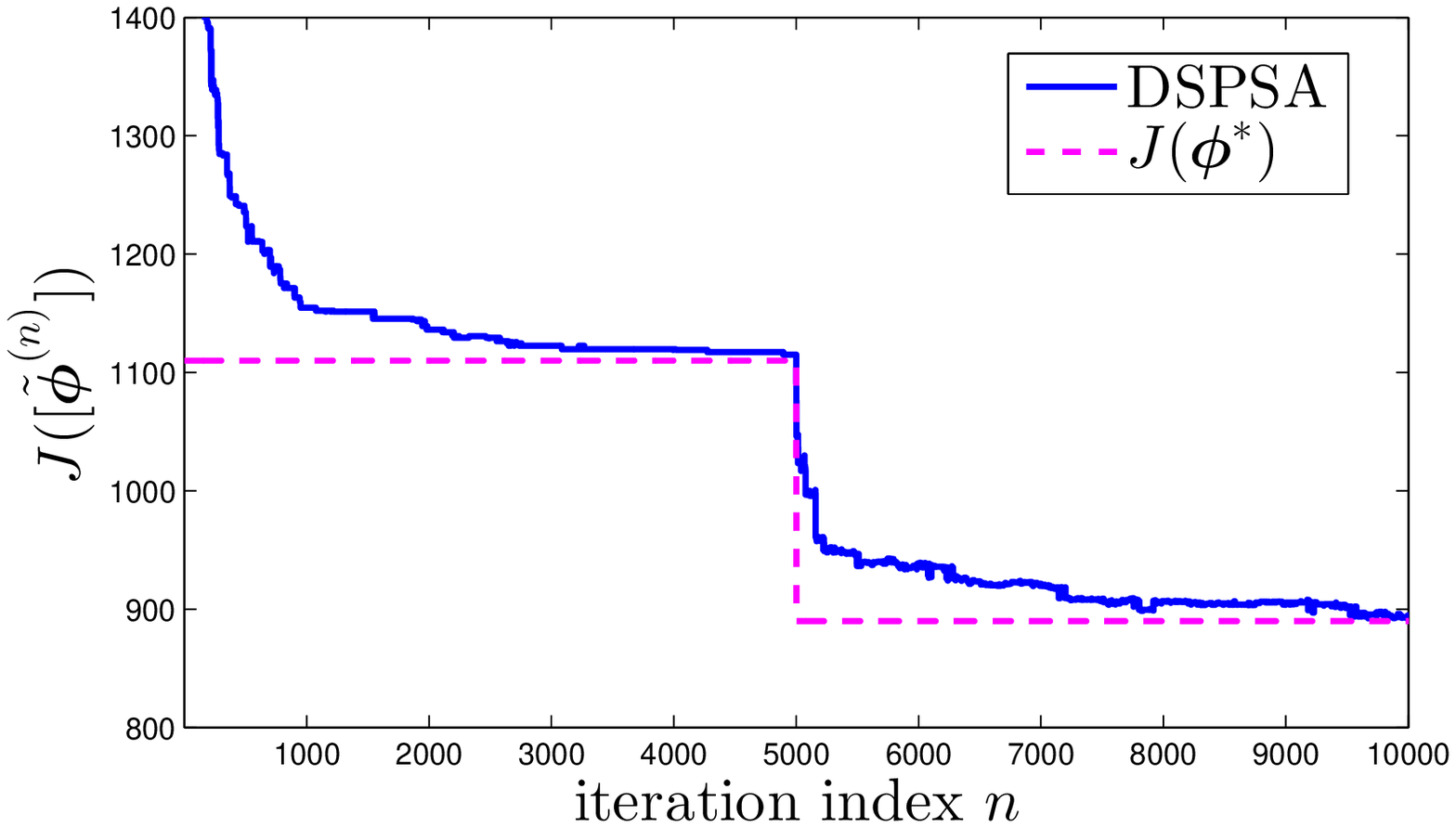}}
        \subfigure[$\tilde{\phi}_{hi}^{(n)}$ when $h=5$ and $i=3$ (right), the $n$th estimation of one tuple of $\Thv^*$]{\includegraphics[height=5cm,width=8cm]{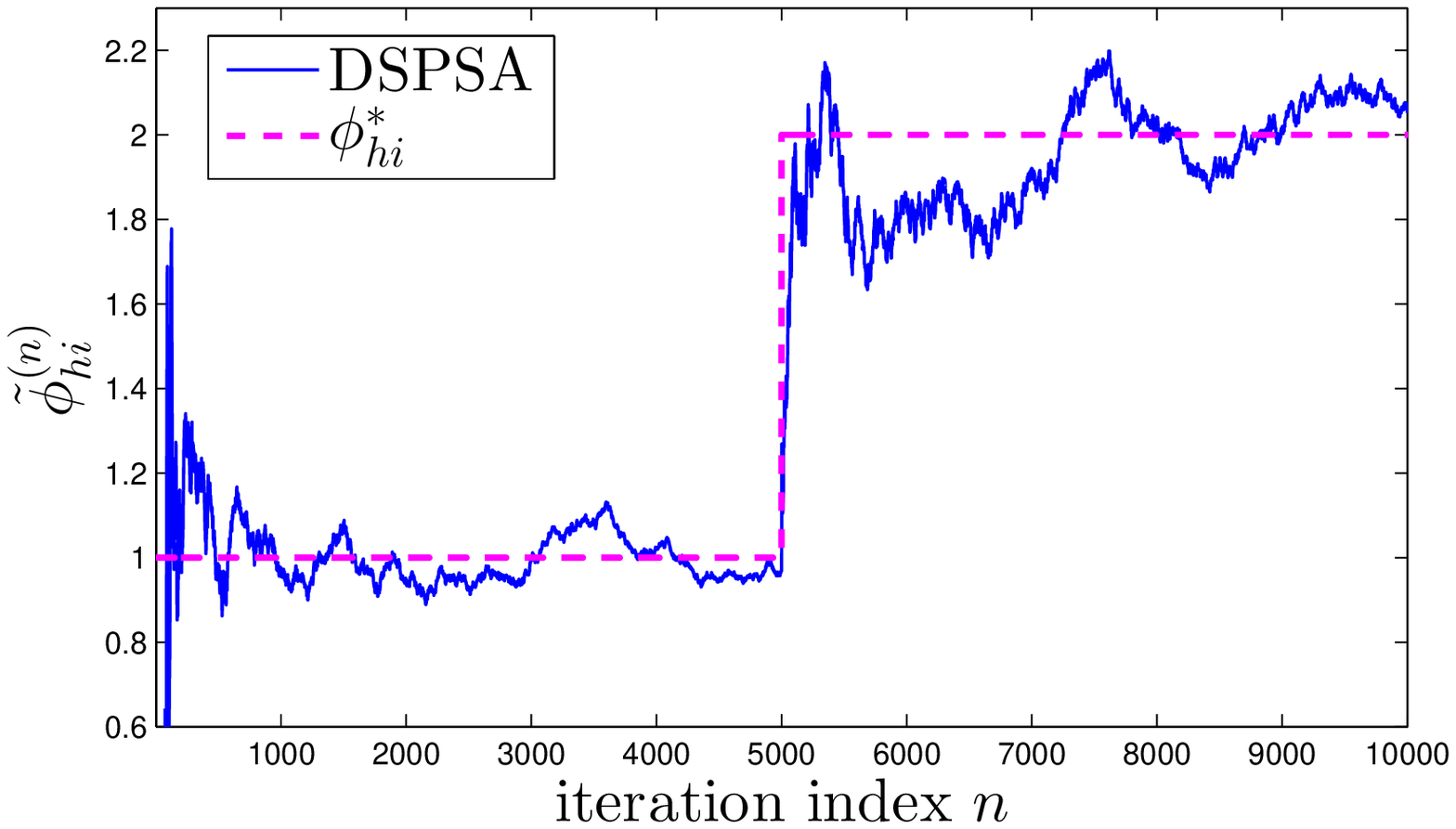}}
	\caption{Convergence performance of DSPSA when we set $w=300$ for the first $5000$ iterations and change to $w=20$ for the second $5000$ iterations. The other parameters are the same as in Fig.~\ref{fig:SPSA2}. }
	\label{fig:SPSAOnLine}
\end{figure*}

\subsubsection{Complexity and Convergence Performance}
\label{sec:Complex}
One advantage of DSPSA is its low complexity: The estimation of $\gv$ in each iteration only requires two simulations of the objective function. It is also proved that the estimation sequence generated by DSPSA is able to converge to the local minimizer for problem~\eqref{eq:objSPSA1} probabilistically\cite{Wang2008SA}. We run experiments to show the convergence performance of DSPSA. We set duration of decision epoch $T_D=10^{-3}$, queue length $\LB=15$, the maximum action $A_m=5$, $f^{(t)}\sim{\text{Pois}(3)}$ and the BER constraint $\bar{P}_e=10^{-3}$. The channel is Rayleigh fading with average SNR being $0\si{\decibel}$ and maximum Doppler shift being $10\text{Hz}$. It is modeled by a $8$-state FSMC. We set discount factor $\beta$ to $0.95$ and the total number of iterations $N$ in DSPSA to $5000$. We first choose $w=100$ and apply DSPSA to search the optimal threshold vector. The convergence performance is shown in Fig.~\ref{fig:SPSA2}. The optimal threshold vector $\Thv^*$ is determined by the optimal policy $\theta^*$ searched by DP. We then set $w=400$ and apply DSPSA again. The results are shown in Fig.~\ref{fig:SPSA3}. It can be seen that DSPSA converges to the optimum in both figures. Based on Figs.~\ref{fig:SPSA2} and \ref{fig:SPSA3}, the convergence speed of DSPSA when $w=400$ is faster than that when $w=100$. We do not have the direct proof of the rate of convergence of DSPSA. But, we provide two possible reasons why DSPSA converges faster with higher value of $w$. One is the shape of the objective function $J$ in the neighbourhood of the local minimizer since a study in \cite{Moulines2011} shows that stochastic steepest descent algorithms converge faster for strongly convex functions than for non-strongly convex functions on average. The other reason is the step size parameters. The step size parameters are important for the convergence performance of stochastic approximation algorithms \cite{SpallPar1998}. In this paper, we follow the suggestions in \cite{SpallPar1998} to set the values of step size parameters. But, there may exist a different set of step size parameters with which the convergence performance when $w=100$ could be improved. After all, Figs.~\ref{fig:SPSA2} and \ref{fig:SPSA3} show that DSPSA is able to approximate an estimator of the optimal queue threshold vector where the value of the objective function is very close to the optimum. One may be interested in studying how to speed up the DSPSA algorithm. But, it is beyond the scope of this paper and could be a proposal of the research work in the future.

The other advantage of DSPSA is that it does not require the full knowledge of MDP. Since DSPSA is a simulation-based algorithm, it can be implemented if only a simulation model is available. Therefore, DSPSA is suitable for real-time applications. Fig.~\ref{fig:SPSAOnLine} shows the convergence performance of DSPSA when we change the value of $w$. We use the same parameters as in Figs.~\ref{fig:SPSA2} and \ref{fig:SPSA3}. We apply DSPSA and change the value of $w$ from $300$ to $20$ at the $5000$th iteration. It can be seen that DSPSA is able to adaptively track the optimum and optimizer accordingly with the changing value of $w$. The results also implies that DSPSA can be combined with model-free learning algorithms for the scheduler to learn the optimal transmission policy in real time.

\section{Conclusion}
We studied the monotonicity of the optimal policy in an MDP modeled cross-layer adaptive $m$-QAM system. It was proved that the optimal policy was always nondecreasing in queue state due to the $\LN$-convexity of DP. By observing the submodularity of DP conditioned on the weight factor in the cost function and the channel statistics, we derived the sufficient conditions for the optimal policy to be nondecreasing in both queue and channel states. We showed that $\LN$-convexity differed from submodularity in that the variation of the resulting optimal policy was not only monotonic but also restricted by a bounded marginal effect. We presented two low complexity algorithms: MPI based on $\LN$-convexity and DSPSA. We showed that MPI based on $\LN$-convexity incurred a much reduced the computational complexity than DP \cite{PutermanMDP1994} and MPI based on submodularity \cite{Djonin2007}. For DSPSA, we ran numerical experiments to show its convergence performance, where we showed that it allowed the decision maker to adaptively trace the optimal policy.

It should be pointed out that the algorithms for finding the monotonic optimal policy in cross-layer adaptive $m$-QAM system is not restricted to MPI and DSPSA. One can use the results in Section~\ref{sec:Mono} to propose more efficient algorithms. For example, one may consider random search or simulated annealing algorithms for solving problem~\eqref{eq:app:obj}. This could be one direction of the research works in the future. In addition, Propositions~\ref{prop:MonoQue} and \ref{prop:MonoChan} are not restricted to expressions of $c_q$ and $c_{tr}$, i.e., they can be utilized to
derive the monotonicity of the optimal policy in other queue-assisted cross-layer transmission control problems. Finally, as discussed in Section~\ref{sec:Complex}, to discuss how to speed up the DSPSA algorithm when it is applied to cross-layer modulation system could be another direction of the research works in the future.

--------------------------------------------------------------------------------------------------------------%

\appendices

\section{}
\label{app:1stDom}
Stochastic dominance is the stochastic ordering that used in decision analysis. It describes a probability distribution is superior to another in terms of the expected outcomes or costs. In this paper, we use the concept of first order stochastic dominance defined blow to show the monotonicity of the optimal policy in channel states.
\begin{definition}[first order stochastic dominance is \cite{Smith2002}] \label{def:1stDom}
Let $\tilde{\rho}(x)$ be a random selection on space $\X$ where $x$ conditions the random selection, then $\tilde{\rho}(x)$ is first order stochastically nondecreasing in $x$ if $\E[u(\tilde{\rho}(x_+))] \geq \E[u(\tilde{\rho}(x_-))]$ for all nondecreasing functions $u$ and $x_+ \geq x_-$.
\end{definition}

\section{}
\label{app:MonoQue2}
Assume $V(\x')$ is nondecreasing in $b'$. It is straightforward to see that $\varphi_o(y,f)$ is nondecreasing in $y$. Since $\min\{[y]^{+}+f,\LB\}$ is nondecreasing in $y$, $\beta V(\min\{[y]^{+}+f,\LB\},h')$ is nondecreasing in $y$. Therefore, $\hat{V}(y,f,h)$ is nondecreasing in $y$. Consider the monotonicity of $Q$ in $b$. Since
    \begin{align}
        &\quad Q(b+1,h,a)-Q(b,h,a)   \nonumber \\
        &=\sum_{h'}P_{hh'} \E_f [ \hat{V}(b-a+1,f,h')-\hat{V}(b-a,f,h') ] \geq{0},
    \end{align}
$Q$ is nondecreasing in $b$.

\section{}
\label{app:MonoQue3}
Since the additions of two $\LN$-convex functions are $\LN$-convex \cite{Zipkin2008}, $Q$ is $\LN$-convex if both $c_{tr}(h,a)$ and $\sum_{h'}P_{hh'} \E_f [\hat{V}(b-a,f,h')]$ are $\LN$-convex in $(b,a)$. Consider the $\LN$-convexity of $c_{tr}$. Since $c_{tr}$ is just a function of $a$, it suffices to show that $c_{tr}$ is $\LN$-convex in $a$. $c_{tr}$ is $\LN$-convex in $a$ since $2^{a}$ is convex in $a$.

Consider the $\LN$-convexity of $\sum_{h'}P_{hh'} \E_f [\hat{V}(b-a,f,h')]$. Since the expectation of $\LN$-convex function is $\LN$-convex \cite{Zipkin2008}, it suffices to show the $\LN$-convexity of $\hat{V}(y,f,h)$ in $(b,a)$. By Definition~\ref{def:Lconvex}, we need to prove that $\psi(b,a,f,h',\zeta)=\hat{V}(b-a,f,h')$ is submodular in $(b,a)$. But, by Definition~\ref{def:submodularity}, $\psi(b,a,f,h',\zeta)$ is submoular in $(b,a)$ since
\begin{align}  \label{eq:app:MonoBH5}
    &\quad \psi(b+1,a,f,h',\zeta)+\psi(b,a+1,f,h',\zeta) - \psi(b,a,f,h',\zeta)-\psi(b+1,a+1,f,h',\zeta)  \nonumber \\
    &=\hat{V}(b-a+1,f,h')+\hat{V}(b-a-1,f,h') - 2\hat{V}(b-a,f,h')\geq{0}
\end{align}
for all $(b,a)$. See the proof in Appendix~\ref{app:MonoQue4} for the last step in \eqref{eq:app:MonoBH5}. Therefore, $\hat{V}(y,f,h)$ is $\LN$-convex in $y$, and $Q$ is $\LN$-convex in $(b,a)$.

\section{}
\label{app:MonoQue4}
Assume that $V(\x')$ is $\LN$-convex in $b'$, $\hat{V}(y,f,h)$ is $\LN$-convex in $y$ because
    \begin{align}
        &\quad \hat{V}(y+1,f,h)+\hat{V}(y-1,f,h)-2\hat{V}(y,f,h)   \nonumber \\
        &=\varphi_o(y+1,f)+\varphi_o(y-1,f)-2\varphi_o(y,f)  + \beta \bigg(V\Big(\min\{[y+1]^{+}+f,\LB\},h\Big)    \nonumber \\
        &\qquad  +V\Big(\min\{[y-1]^{+}+f,\LB\},h\Big)   -  2V\Big(\min\{[y]^{+}+f,\LB\},h\Big) \bigg)   \nonumber  \\
        &=\begin{cases}
                0 & y<0 \\
                w \geq{0} & y=0 \\
                \beta \Big( V(\tilde{y}+1,h) + V(\tilde{y}-1,h) - 2V(\tilde{y},h) \Big) \geq{0} & 0<y<\LB-f  \\
                w+\beta \Big( V(\LB-1,h) - V(\LB,h) \Big)  & y=\LB-f   \\
                0 & \LB-f<y\leq{\LB}
          \end{cases},  \nonumber
    \end{align}
where $\tilde{y}=y+f$. Let $a_{b}^*=\arg\min_{a}Q(\LB,h,a)$ and $a_{\LB-1}^*=\arg\min_{a}Q(\LB-1,h,a)$. We have
    \begin{align}  \label{eq:app:MonoB1}
        &\quad V(b,h)-V(b+1,h)    \nonumber \\
        &=Q(b,h,a_{b}^*)-Q(\LB,h,a_{b+1}^*)   \nonumber \\
        &\geq Q(b,h,a_{b}^*)-Q(b+1,h,a_{b}^*)    \nonumber \\
        &= \sum_{h'}P_{hh'} \E_f \Big[ \hat{V}(b-a,f,h)-\hat{V}(b+1-a,f,h) \Big]  \geq -w.
    \end{align}
So $w+\beta\Big(V(\LB-1,h)-V(\LB,h)\Big)\geq w-w\beta \geq 0$. Therefore, $\hat{V}$ is $\LN$-convex in $y$.

\section{}
\label{app:MonoQueChan}
Assume that $V(\x')$ is submodular in $\x'=(b',h')$. $Q$ is submodular in $(b,h)$ because
    \begin{align}
        &\quad Q(b,h+1,a)+Q(b,h,a+1)-Q(b,h,a) - Q(b,h+1,a+1)                                                          \nonumber \\
        &=c_{tr}(h+1,a)+c_{tr}(h,a+1)-c_{tr}(h,a) -c_{tr}(h+1,a+1)     \nonumber \\
        &\qquad +\sum_{(h+1)'}P_{(h+1)(h+1)'} \E_f \Big[ \hat{V}(b-a,f,(h+1)') -\hat{V}(b-a-1f,(h+1)') \Big]                   \nonumber \\
        &\qquad + \sum_{h'}P_{hh'} \E_f \Big[ \hat{V}(b-a-1,f,h') - \hat{V}(b-a,f,h') \Big]                                                    \nonumber \\
        &\geq w+\sum_{h'}P_{hh'} \E_f \Big[ \hat{V}(b-a-1,f,h') -\hat{V}(b-a,f,h') \Big]         \label{eq:app:MonoBH1} \\
        &\geq w-w  \geq{0},     \label{eq:app:MonoBH2}
    \end{align}
    \begin{align}
        &\quad Q(b+1,h,a)+Q(b,h+1,a)-Q(b,h,a)  - Q(b+1,h+1,a)                                                          \nonumber \\
        &=\sum_{(h+1)'}P_{(h+1)(h+1)'} \E_f \Big[ V(\min\{[b-a]^{+}+f,\LB\},(h+1)')    \nonumber \\
         &\qquad - V(\min\{[b-a+1]^{+}+f,\LB\},(h+1)') -\sum_{h'}P_{hh'} \E_f \Big[ V(\min\{[b-a]^{+}+f,\LB\},h')     \nonumber \\
         &\qquad  - V(\min\{[b-a+1]^{+}+f,\LB\},h')   \big] \geq{0} \label{eq:app:MonoBH3}
    \end{align}
and
    \begin{align}
        &\quad Q(b+1,h,a)+Q(b,h,a+1)-Q(b,h,a) -Q(b+1,h,a+1) \geq{0}.  \label{eq:app:MonoBH4}
    \end{align}
Here, \eqref{eq:app:MonoBH1} is because $\hat{V}$ is nondecreasing in $y$ as proved in Appendix~\ref{app:MonoQue2}. \eqref{eq:app:MonoBH2} is because $\sum_{h'}P_{hh'} \E_f \Big[ \hat{V}(b-a-1,f,h')-\hat{V}(b-a,f,h') \Big] \geq{-w}$ as proved in \eqref{eq:app:MonoB1}. \eqref{eq:app:MonoBH3} is because of the submodularity of $V(\x')$ in $\x'=(b',h')$ and first order stochastic monotonicity of $P_{hh'}$ in $h$. \eqref{eq:app:MonoBH4} is due to the $\LN$-convexity of $Q$ in $(b,a)$ as shown in Appendix~\ref{app:MonoQue3}.

\newpage

\bibliographystyle{elsarticle-num}
\bibliography{AMCbib}

\end{document}